\documentclass{article}

\usepackage[preprint]{neurips_2026}

\usepackage[utf8]{inputenc}
\usepackage[T1]{fontenc}
\usepackage{hyperref}
\usepackage{url}
\usepackage{booktabs}
\usepackage{amsfonts}
\usepackage{amsmath,amssymb,amsopn}
\usepackage{nicefrac}
\usepackage{microtype}
\usepackage{pgfplots}
\usetikzlibrary{pgfplots.groupplots}
\usepackage{xcolor}
\usepackage{graphicx}
\usepackage{multicol}
\usepackage{multirow}
\usepackage{caption}
\usepackage{verbatim}
\usepackage[utf8]{inputenc}
\usepackage[T1]{fontenc}
\usepackage{url}
\usepackage{graphicx}
\usepackage{amsfonts}
\usepackage{amsmath}
\usepackage{amssymb}
\usepackage{pgfplots}
\usepackage{subcaption}
\usepackage{booktabs}
\usepackage{multirow}
\usepackage{xcolor}
\usepackage{tocloft}
\usepackage{pgfgantt}
\usepackage{caption}
\usepackage{float}
\usepackage{setspace}
\usepackage[most]{tcolorbox}
\usepackage{enumitem}
 \usepackage[misc]{ifsym}

\definecolor{ForestGreen}{RGB}{34,139,34}

\title{Does Synthetic Layered Design Data Benefit Layered Design Decomposition?}

\author{%
Kam Man Wu\textsuperscript{1}\thanks{Equal contribution.} \hspace{0.2cm}
Haolin Yang\textsuperscript{1}\footnotemark[1] \hspace{0.2cm}
Qingyu Chen\textsuperscript{2} \hspace{0.15cm}
Yihu Tang\textsuperscript{2} \hspace{0.15cm} \\
\textbf{Jingye Chen}\textsuperscript{1}\thanks{Project lead.}\;\thanks{Corresponding authors ($^{\textrm{\Letter}}$ jchenha@connect.ust.hk\;\;$^{\textrm{\Letter}}$ cqf@ust.hk).}\hspace{0.3cm}
\textbf{Qifeng Chen}\textsuperscript{1}\footnotemark[3] \\[0.15cm]
\textsuperscript{1}HKUST \qquad
\textsuperscript{2}Webank \\[0.1cm]
}

\begin{document}

\maketitle

\begin{abstract}

Recent advances in image generation have made it easy to produce high-quality images. However, these outputs are inherently flattened, entangling foreground elements, background, and text within a fixed canvas. As a result, flexible post-generation editing remains challenging, revealing a clear last-mile gap toward practical usability.
Existing approaches either rely on scarce proprietary layered assets or construct partially synthetic data from limited structural priors. However, both strategies face fundamental challenges in scalability.
In this work, we investigate whether pure synthetic layered data can improve graphic design decomposition. We make the simplifying assumption that, in graphic design, effective decomposition does not require modeling inter-layer dependencies as precisely as in natural-image composition, since design elements are often intentionally arranged as modular and semantically separable components.
Concretely, we conduct a data-centric study based on CLD baseline, which is a state-of-the-art layer decomposition framework. Based on the baseline, we construct our own synthetic dataset, SynLayers, generate textual supervision using vision language models, and automate inference inputs with VLM-predicted bounding boxes. Our study reveals three key findings: (1) even training with purely synthetic data can outperform non-scalable alternatives such as the widely used PrismLayersPro dataset, demonstrating its viability as a scalable and effective substitute; (2) performance consistently improves with increased training data scale, while gains begin to saturate at around 50K samples; and (3) synthetic data enables balanced control over layer-count distributions, avoiding the severe layer-count imbalance commonly observed in real-world datasets. We hope this data-centric study encourages broader adoption of synthetic data as a practical foundation for layered design editing systems. We make the code publicly available at 
\href{https://github.com/YangHaolin0526/SynLayers}{GitHub}.

\end{abstract}

\section{Introduction}

Graphic design plays a central role in modern content creation, including advertising, social media campaigns, presentations, and digital publishing. As visual generative models continue to mature, they have drawn increasing attention to the last mile of practical editing and controllability ~\citep{zhang2025creatiposter,wei2025postercopilot,zhang2025creatidesign}.
In practice, designers compose content in layers using tools such as Adobe Photoshop or PowerPoint, but the exported result is typically a single raster image in which foreground elements, background, and text are flattened together~\citep{suzuki2025layerd}. Consequently, subsequent edits often require manual effort, while generative editing \citep{hertz2022prompt,meng2021sdedit,jia2024designedit,brack2024ledits++} can be unstable and may fail to preserve the geometry and semantics of unchanged regions, even for simple operations such as background flipping or text resizing~\citep{brooks2023instructpix2pix,couairon2022diffedit,simsar2025lime}.
These limitations motivate graphic design decomposition, which aims to recover editable RGBA layers from raster images for reliable and fine-grained control.

In this case, graphic design decomposition can be viewed as the ``last mile'' that bridges image generation and user-editable outputs, and has recently attracted growing attention from the research community ~\citep{suzuki2025layerd,yin2025qwen,nie2025decomposition}. Existing approaches typically rely on limited amounts of real, non-synthetic graphic design data (\textit{e.g.}, Crello with around 20K samples) or on partially synthetic datasets such as PrismLayersPro with 20K samples, in which layout boxes are derived from real designs. At the same time, visual assets are generated by image models. However, these dataset sizes remain orders of magnitude smaller than those commonly used in modern generative modeling—for instance, text-to-image models are often trained on billions of samples (\textit{e.g.}, LAION-5B ~\citep{schuhmann2022laion}). This contradicts established scaling laws ~\citep{kaplan2020scaling,hoffmann2022training}, and models trained on limited data often lack robustness and generalization. Therefore, a promising alternative is to leverage fully synthetic data, an approach that has demonstrated effectiveness across many domains~\citep{fan2024scaling,man2022review}. Yet its potential for layer decomposition remains largely unexplored. In this work, we aim to fill this gap by investigating whether purely synthetic layer data can effectively benefit graphic design decomposition.

\begin{figure}[t]
    \centering
    \includegraphics[width=1\linewidth]{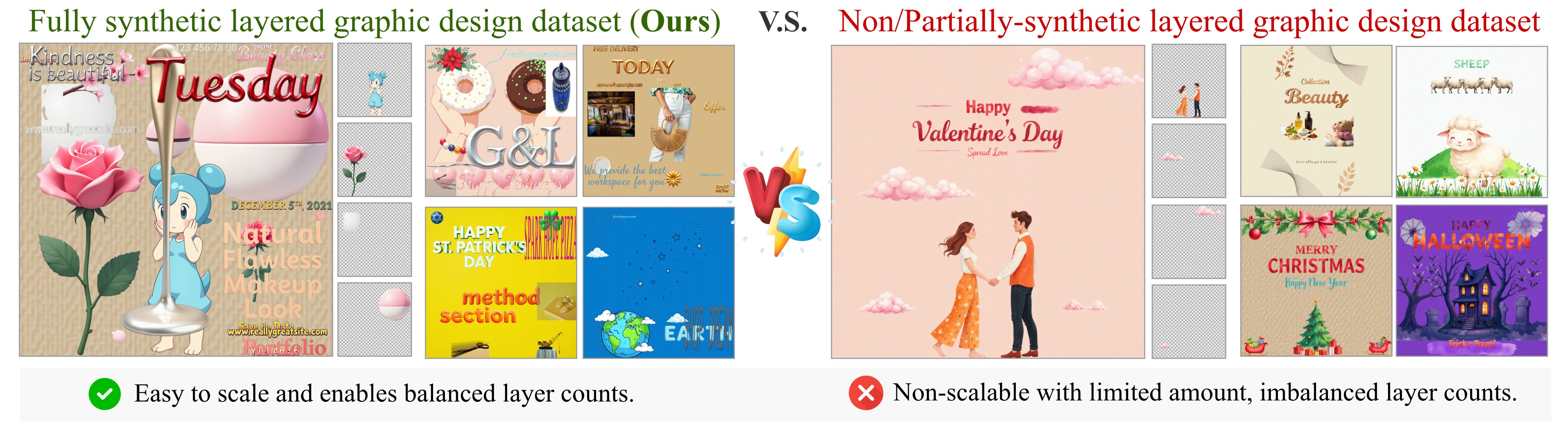}
    \caption{Comparison between fully synthetic graphic-design data from SynLayers and non/partially-synthetic counterparts. In this paper, we explore whether synthetic layer data can help train effective layer decomposition models. We show only four layers from the decomposition results due to limited space.}
    \label{fig:teaser}
\end{figure}

To answer this question, we propose SynLayers. We carefully design a synthetic data pipeline in which graphic design layers are composed from multiple sources, and conduct a comprehensive data-centric study. Building on the representative state-of-the-art layer decomposition method, CLD~\citep{liu2025controllable}, we systematically investigate the role of synthetic data in this task. Specifically, we design layer composition rules to construct diverse synthetic samples. Figure \ref{fig:teaser} shows the advantages of SynLayers. We further fine-tune a VLM to generate high-quality captions for our constructed synthetic data, providing additional semantic supervision for layer decomposition. Through extensive experiments, our data-centric study aims to answer the following questions:

\begin{itemize}[leftmargin=1em]
    \item \textbf{\textit{Can synthetic data serve as a viable alternative to non-/partially synthetic datasets?}} Yes. With the same amount of training data (18K samples), synthetic data achieves comparable or better performance on most metrics, with improvements in both layer PSNR (+1.01) and composite PSNR (+0.83), indicating stronger reconstruction quality at both the layer and image levels.
    \item \textbf{\textit{Does synthetic data follow scaling laws, i.e., does performance improve consistently with increased data size?}} 
    No. Performance does not improve monotonically with more synthetic data. For example, the best results are achieved at medium scales, with Layer FID reaching 5.97 at 20K and Composite FID reaching 10.35 at 30K.
    \item \textbf{\textit{Does synthetic data improve training stability regarding distribution balance?}} Yes. Synthetic data yields more robust performance across different layer counts. It consistently outperforms the baseline across different layer-count settings, with more gains in challenging cases. For example, when the number of layers is between 13 and 35, Mask IoU improves from 0.901 to 0.910, and Composite PSNR improves from 29.48 to 30.25. 
\end{itemize}

We hope this study demonstrates that synthetic data is a practical and scalable option for training layer-decomposition models and encourages its use in building more reliable and controllable graphic design editing systems.

\section{Related Works}

\subsection{Layered Image Decomposition}
Layered image decomposition separates raster images into editable RGBA layers, enabling flexible reconstruction and modification. Several research works \cite{liu2025magicquillv2,liu2025omnipsd,lungu2026lade,kang2025layeringdiff,huang2025dreamlayer,zhang2023text2layer,zhang2024transparent,dalva2024layerfusion,huang2024layerdiff} focus on this problem. For example, LayerD~\citep{suzuki2025layerd} employs iterative matting and inpainting but offers limited user control, while Accordion~\citep{chen2025rethinking} utilizes VLMs and SAM~\citep{kirillov2023segment} in a top-down pipeline to recover layers from flattened AI-generated designs. Bottom-up and layout-aware approaches include COLE~\citep{jia2023cole} and Open-COLE~\citep{inoue2024opencole}, which cascade LLMs with diffusion models, and Composition-Aware Graphic LayoutGAN~\citep{zhou2022composition}, which learns structural priors for visual-textual arrangements. Recently, diffusion-based models such as Qwen-Image-Layered~\citep{yin2025qwen} have introduced variable-length disentanglement via RGBA-VAE and VLD-MMDiT. However, it suffers from high inference latency due to the heavy diffusion design. Notably, CLD~\citep{liu2025controllable}, which is built on top of ART \cite{pu2025art}, introduces LD-DiT and MLCA components for bounding-box-guided decomposition, offering an efficient alternative to heavier diffusion design. Given the trade-off between controllability and computational efficiency, we adopt CLD as our basic framework. 

\subsection{Datasets for Multi-Layer Graphic Design Decomposition}

Multi-layer datasets are limited because clean RGBA layers are usually stored in proprietary design files rather than in publicly available raster image datasets. Existing resources such as MuLan~\citep{tudosiu2024mulan} and MAGICK~\citep{burgert2024magick} provide some layer annotations or simple composites, but they lack the complexity and rich text content needed for graphic design tasks. Within the graphic design domain, existing resources are constrained by scale and synthesis depth. CanvasVAE~\citep{yamaguchi2021canvasvae} releases a layered graphic dataset, Crello, while PrismLayersPro~\citep{chen2025prismlayers} provides a high-quality dataset of 20K samples with transparent layers. Recently, LICA~\citep{hirsch2026lica} introduced a layered image composition annotation strategy to support graphic design research. As shown in Table \ref{tab:comparison}, we present a clear comparison between SynLayers and existing alternatives to explore the potential of synthetic data.

\begin{table}[t]
\caption{Comparison with existing multi-layer graphic design datasets. Our work fills the gap by investigating purely synthetic data. $^{*}$SynLayers can scale to much larger datasets, and we use 500K samples for analysis.}
\label{tab:comparison}
\centering
\begin{tabular}{lcccc}
\toprule
Dataset & \#Samples & \#Layers & Source & Scalable \\
\midrule
Crello~\citep{yamaguchi2021canvasvae} & 23K & $\leq 50$ & Real & No \\
PrismLayersPro~\citep{chen2025prismlayers} & 20K & $\leq 52$ & Partially Synthetic  & No \\
\midrule
\textbf{SynLayers (Ours)} & 500K$^{*}$ & $\leq 52$ & Fully Synthetic & Yes \\
\bottomrule
\end{tabular}
\end{table}

\subsection{Synthetic Data for Vision Tasks}

Synthetic data has become a cornerstone of modern computer vision pipelines, particularly for mitigating data scarcity ~\citep{delussu2024synthetic,mumuni2024survey,nikolenko2021synthetic,jordon2022synthetic,westerski2024synthetic,nowruzi2019much,vanherle2022analysis}. The limitations of existing layered design datasets reflect broader challenges where curated real-world supervision is expensive or difficult to obtain. In such scenarios, synthetic data generation has proven effective across diverse domains. For example, in OCR tasks, synthetic text generation has long been employed to supplement training data, with early works utilizing millions of synthetic samples to improve recognition robustness~\citep{jaderberg2014synthetic}. In the generative era, models like Qwen-Image utilize synthetic visuals, including slides and banners, to address the lack of text-rich real-world training data~\citep{wu2025qwen}. Similarly, studies have demonstrated that synthetic images can achieve parity with real data in recognition tasks~\citep{he2022synthetic}, while in 3D vision, synthetic assets facilitate training reconstruction models such as LRM-Zero~\citep{xie2024lrm}. Inspired by these successes, we utilize SynLayers to investigate the effectiveness of synthetic data for graphic design decomposition.

\section{Methodology}

\begin{figure}[t]
    \centering
    \includegraphics[width=\linewidth]{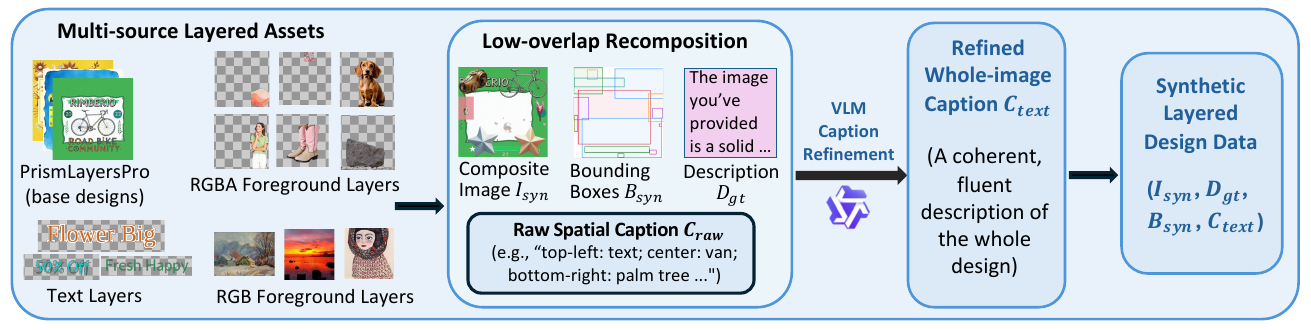}
    \caption{Overview of construction of SynLayers. Multi-source assets, including base designs, RGBA/RGB foregrounds, and text layers, are recombined with a low-overlap algorithm to generate composite images, spatial bounding boxes, and raw layout descriptions. A VLM refines these into coherent whole-image captions. The output is a fully synthetic layered dataset comprising composite images, ground-truth layer boxes, and structured captions, which provides complete supervision for decomposition training.}
    \label{fig:pipeline}
\end{figure}

This section is organized as follows. Section~\ref{sec:cld_problem} introduces CLD as the preliminary knowledge. Section~\ref{sec:dataset} describes the strategy of processing multi-source data and caption construction. Section~\ref{sec:vlm_detection} trains a VLM for layout detection for automatic decomposition. In Figure \ref{fig:pipeline}
We illustrate the construction procedure of SynLayers.

\subsection{Preliminary Knowledge of CLD}

\label{sec:cld_problem}
Our work uses CLD as the base decomposition architecture, which we describe in detail here. Given a raster image $I \in \mathbb{R}^{H \times W \times 3}$, the objective is to generate a set of RGBA layers $\mathcal{D} = \{D_0, D_1, \ldots, D_{N-1}\}$, where $D_0$ denotes the composited image, $D_1$ represents the background layer, and $D_2$ to $D_{N-1}$ correspond to the foreground alpha layers. Following CLD, the decomposition is conditioned on a whole-image caption $C_\text{text}$ and a set of bounding boxes $\mathcal{B} = \{B_\text{comp}, B_0, B_1, \ldots, B_{N-1}\}$. Here, $B_\text{comp}$ and $B_0$ correspond to the composite and background images, both of which span the full image region with coordinates $[0, 0, W, H]$, while the remaining boxes ${B_1, \ldots, B_{N-1}}$ denote foreground regions.
For any box $B_i = (x_i^l, y_i^l, x_i^r, y_i^r)$, each coordinate is adjusted to be divisible by 16.

CLD adopts a crop-then-denoise pipeline. Instead of processing the entire image, it first crops features from each layer's bounding box and then performs joint denoising across all cropped regions. This design restricts computation to the spatial extent of valid layers, making the model more efficient and more scalable to varying numbers of layers than Qwen-Image-Layered~\citep{yin2025qwen}. Finally, a transparent decoder maps the denoised representations back to RGBA layers.


\subsection{Multi-Source Synthetic Dataset Construction}
\label{sec:dataset}

\begin{figure}[t]
    \centering
    \includegraphics[width=1\linewidth]{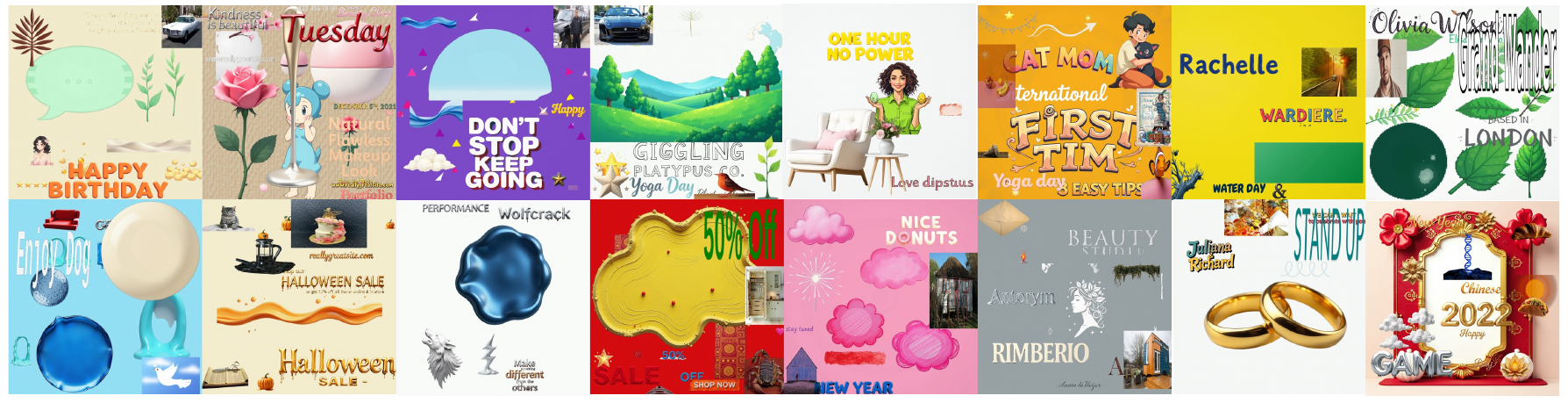}
    \caption{Our samples illustrate the strength of SynLayers, which are fully synthetic, with high-quality RGBA layers, and ensure the layout diversity.}
    \label{fig:comparison_2}
\end{figure}

High-quality graphic design data with per-layer RGBA supervision remains scarce. To address this limitation, we construct SynLayers, a fully synthetic training set generated via a multi-source layer-composition pipeline that systematically assembles diverse elements into coherent designs. Figure \ref{fig:comparison_2} shows the samples selected from SynLayers. Each sample is generated through the following procedure:
\begin{itemize}[leftmargin=1em]
\item \textbf{Base layout construction.} We initialize from a reference design, preserving its background and a subset of foreground elements to maintain structural consistency.

\item \textbf{Cross-source layer integration.} We sample additional designs and incorporate selected foreground elements, increasing compositional diversity while preserving design plausibility.

\item \textbf{Auxiliary element augmentation.} Additional elements are introduced following predefined probabilities:
\begin{itemize}
    \item LAION image crops (probability $0.60$, relative scale in $[0.3, 0.4]$ of the canvas),
    \item Rendered text layers (probability $0.35$, relative scale in $[0.6, 0.8]$ of the canvas),
    \item AlphaVAE foreground objects (up to 3 instances per sample, relative scale in $[0.25, 0.40]$ of the canvas)~\citep{schuhmann2022laion,wang2025alphavae}.
\end{itemize}

\item \textbf{Spatial arrangement.} Layer placement is determined via candidate sampling to reduce overlap and ensure spatial coherence.

\end{itemize}

Full details of the generation pipeline, including size configurations, are provided in Appendix~\ref{app:method_details}.

\paragraph{Layer composition and placement.}
To place each layer, we aim to minimize spatial overlap with existing elements. For a layer with size $(w_j, h_j)$, we sample a set of candidate positions $\mathcal{S}_j$ and select the position that minimizes the normalized overlap:
\[
(x_0^*, y_0^*) = \arg\min_{(x_0, y_0) \in \mathcal{S}_j} 
\frac{1}{\mathrm{Area}(B_j)} 
\sum_{B_k \in \mathcal{B}_\text{occ}} \mathrm{Area}(B_j \cap B_k),
\]
where $B_j = [x_0, y_0, x_0 + w_j, y_0 + h_j]$ denotes the bounding box of the candidate placement, and $\mathcal{B}_\text{occ}$ is the set of previously placed boxes. If a zero-overlap candidate exists, it is selected. Otherwise, the candidate minimizing overlap is chosen. This procedure preserves the aspect ratio of each layer while promoting spatial separation and layout diversity.

\paragraph{Caption construction and refinement.}
\label{sec:caption}
We divide the canvas into a $3 \times 3$ grid, resulting in nine coarse spatial regions (\textit{e.g.}, top-left, top, and bottom-right). Each layer is assigned to one of these regions based on the center of its bounding box. We construct an initial caption by traversing the grid in reading order and describing the content in each region. For each layer, we combine its spatial location with a source-specific description, which may come from PrismLayersPro metadata, LAION image-text pairs, rendered text layers, or AlphaVAE layers. The resulting caption consists of spatial phrases paired with corresponding content descriptions. To improve fluency and coherence, we further refine this draft using a VLM conditioned on the composite image. The final caption is used as text supervision for both CLD training and the downstream detector (see Appendix~\ref{app:caption_refinement}).

\subsection{VLM-Guided Automated Input Generation}
\label{sec:vlm_detection}
CLD requires foreground boxes and a global caption as inputs during inference. Instead of relying on dataset annotations as in prior work~\citep{liu2025controllable}, we train a single VLM to jointly predict both from the raster image in one forward pass, i.e., $f_\psi: I \mapsto (\hat{C}_\text{text}, \hat{\mathcal{B}}_u)$. We build the training data from the SynLayers by pairing each image with its caption and layer boxes, and formatting them as instruction-following training samples. We then fine-tune Qwen3-VL-8B-Instruct~\citep{qwen3vl} with LoRA~\citep{hu2022lora} to predict both captions and bounding boxes. At inference time, the VLM outputs a JSON object containing the image caption and predicted boxes. We further add the composite and background boxes, and quantize all boxes to the 16-pixel grid required by CLD. See Appendix~\ref{app:vlm_illustration} and~\ref{app:detector_training} for details.
This creates a pipeline where the structured outputs from the upstream model directly condition the downstream model~\citep{yang2025marssqlmultiagentreinforcementlearning}.

\section{Experimental Results}

All models are fine-tuned from FLUX.1[dev]~\citep{labs2024flux} using LoRA with rank 64 on LD-DiT and MLCA. We follow CLD~\citep{liu2025controllable} and keep the denoising objective unchanged. To ensure a fair comparison, all models use the same training settings, so performance differences primarily reflect the impact of the training data. Training details are provided in Appendix~\ref{app:decomposition_training}. Detailed comparison between the ground truth and SynLayers' output is in Appendix~\ref{app:gt_vs_SynLayers}.

\subsection{Qualitative Results}

\begin{figure}[t]
    \centering
    \includegraphics[width=\linewidth]{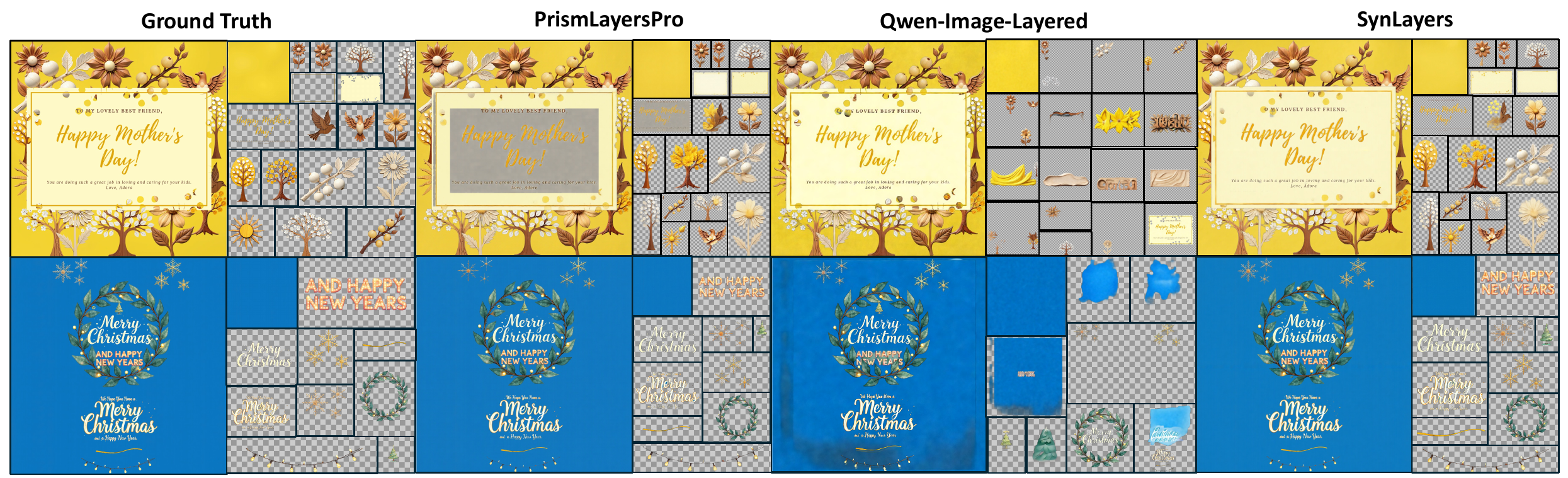}
    \caption{Qualitative comparison between the original PrismLayersPro-trained CLD baseline, our SynLayers-trained model, Qwen-Image-Layered, and the ground-truth decomposition. Across examples, the SynLayers-trained model produces sharper typography, cleaner object boundaries, and layer layouts that more closely match the target decomposition.}
    \label{fig:layer_decomposition2}
\end{figure}

Figure~\ref{fig:layer_decomposition2} presents qualitative comparisons between methods. Both CLD-based models (trained in SynLayers or PrismLayersPro) substantially outperform Qwen-Image-Layered~\citep{yin2025qwen} in the testset, which produces misaligned decompositions with natural layer structures. This stems from Qwen-Image-Layered's requirement for a predefined layer count and its tendency to over-segment semantic elements into fragmented parts. Even when instructed with the correct layer count, it fails to achieve meaningful decomposition, highlighting the advantage of our VLM-detector pipeline, which automatically predicts layer counts and semantic descriptions.

\begin{figure}[!htbp]
    \centering
    \includegraphics[width=\linewidth]{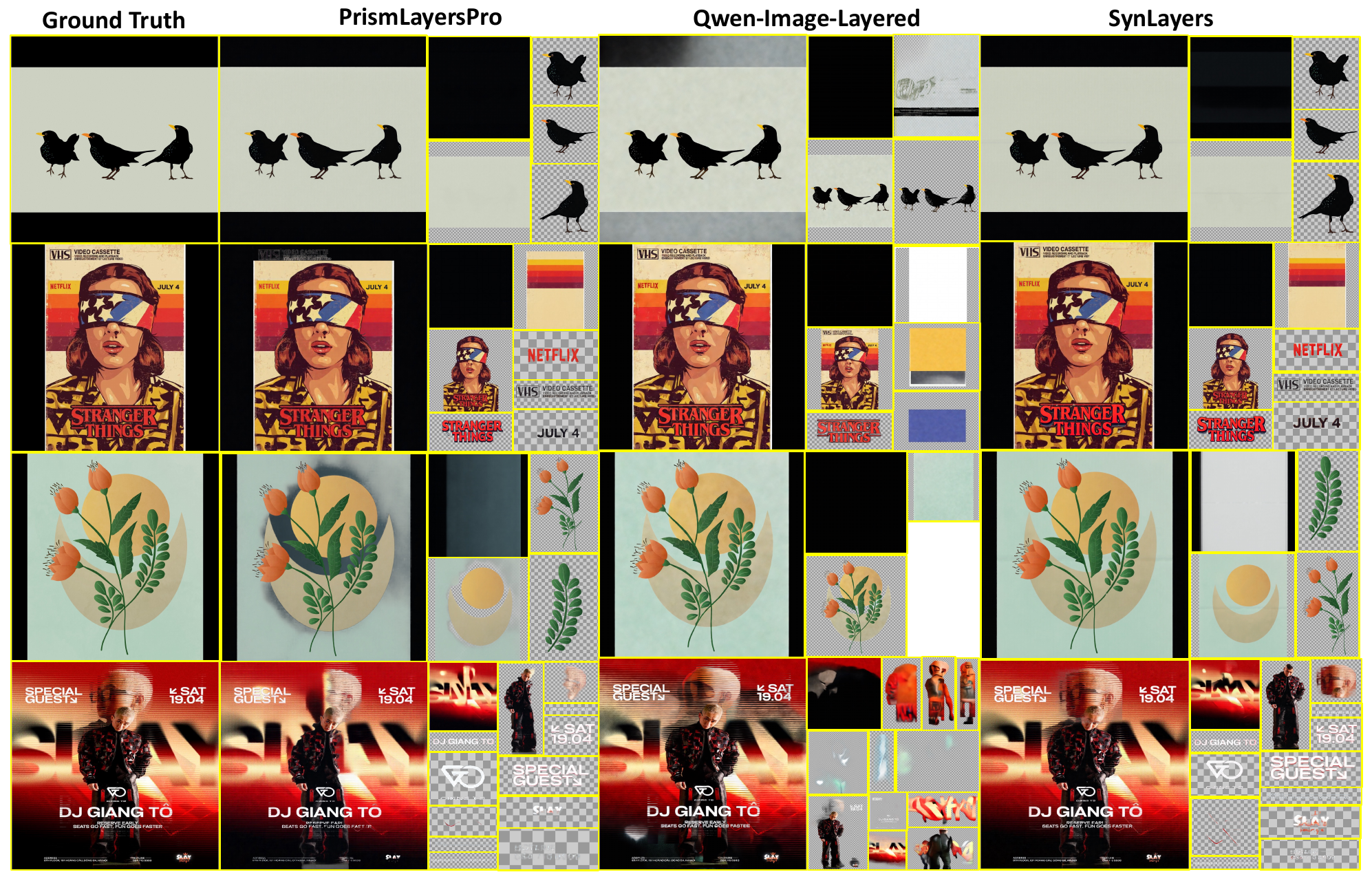}
    \caption{Qualitative comparison on the out-of-distribution real-world dataset. From left to right: Ground Truth composite image, the original CLD baseline, our SynLayers-trained model, and Qwen-Image-Layered. Driven entirely by the Qwen3-VL detector's output, our method yields cleaner semantic separations, sharper details, and higher composite fidelity compared to the baseline and external references.}
    \label{fig:real_comparison}
\end{figure}

Comparing our SynLayers-trained model with the PrismLayersPro baseline reveals clear quality improvements. Our model produces sharper object boundaries with fewer visual artifacts (Fig.~\ref{fig:layer_decomposition2}, row 1, col. 2 vs. 4), such as fewer blue patches inside letter ``e''. Text segmentation is enhanced: the ``Merry Christmas'' text layer (row 2) shows cleaner character separation, and small fonts such as ``We Hope You Have a'' exhibit sharper character appearance and clearer visual effect compared to baseline. These improvements arise from the diverse caption layers in our SynLayers, which better prepare the model for text decomposition. In real-world samples, as Figure~\ref{fig:real_comparison} shows, SynLayers shows better performance than the PrismLayersPro baseline and Qwen-Image-Layered. For example, when evaluating the fourth row, our SynLayers-trained model achieves the best visual clarity when decomposing the human head layer compared to Qwen-Image-Layered and PrismLayersPro baseline. Overall, SynLayers-trained model achieves cleaner visual effects, more accurate object boundaries, and layer layouts that more closely match ground-truth decompositions, demonstrating the effectiveness of SynLayers.

\subsection{Quantitative Results}
Table~\ref{tab:scale_ablation} quantifies how the training scale affects the quality of the decomposition. In a comparison of the same dataset size between PrismLayersPro of 18K samples and SynLayers of 18k samples. SynLayers-trained model outperforms the PrismLayersPro baseline across key metrics. For example, Layer PSNR from $26.22$ to $27.23$ and Composite SSIM from $0.944$ to $0.950$. This result confirms that purely synthetic layers are a viable and data-efficient substitute. Our 1k to 500k dataset size scaling analysis reveals that the training effect gains peak at medium sizes: Layer FID reaches its minimum $5.97$ at 20K, while Composite FID improves to $10.35$ at 30K. Beyond this range, performance stabilizes with diminishing returns, indicating that SynLayers-trained model achieves strong reconstruction quality without requiring massive-scale training.

Table~\ref{tab:qwen_cld_qual} provides a direct reference comparison to Qwen-Image-Layered and the PrismLayersPro baseline. Qwen-Image-Layered's poor layer-wise scores, like PSNR $13.80$ and FID $198.34$, are due to its requirement for a fixed layer count and a tendency to over-segment semantic elements, which misaligns with open-ended editing needs. Although composite reconstruction offers a fairer metric by mitigating fragmentation artifacts, Qwen-Image-Layered still lags behind both CLD and our model. Our 20K checkpoint consistently improves layer-wise and composite fidelity over the baseline. For example, decreasing Layer FID from $6.62$ to $5.97$ and Composite FID from $12.50$ to $12.00$, confirming that synthetic training preserves CLD's decomposition behavior while enhancing reconstruction quality and practical editability.

\begin{table}[t]
\centering
\caption{Quantitative comparison across training-set scales. The table includes the PrismLayersPro reference and a matched-size SynLayers 18K subset, enabling direct comparison at the same nominal data budget.}
\label{tab:scale_ablation}
\resizebox{\textwidth}{!}{
\begin{tabular}{l|c|ccc|cccc|ccc}
\toprule
\multirow{2}{*}{\textbf{Training Set}} & \multirow{2}{*}{\textbf{\# Samples}} & \multicolumn{3}{c|}{\textbf{Layer-wise Quality}} & \multicolumn{4}{c|}{\textbf{Layout Fidelity}} & \multicolumn{3}{c}{\textbf{Composite Quality}} \\
\cline{3-12}\noalign{\vskip 3.5pt}%
 & & PSNR $\uparrow$ & SSIM $\uparrow$ & FID $\downarrow$ & IoU $\uparrow$ & Prec.\ $\uparrow$ & Recall $\uparrow$ & F1 $\uparrow$ & PSNR $\uparrow$ & SSIM $\uparrow$ & FID $\downarrow$ \\
\midrule
PrismLayersPro & 18K & 26.22 & 0.865 & 6.62 & 0.910 & 0.961 & 0.932 & 0.948 & 30.52 & 0.944 & 12.50 \\
\midrule
\multirow{13}{*}{\textbf{SynLayers (Ours)}} & 1K & 25.14 & 0.845 & 6.84 & 0.892 & 0.940 & 0.943 & 0.936 & 28.85 & 0.932 & 17.81 \\
 & 2K & 26.20 & 0.864 & 6.59 & 0.900 & 0.936 & 0.956 & 0.940 & 30.25 & 0.940 & 14.28 \\
 & 3K & 26.35 & 0.870 & 6.14 & 0.912 & 0.944 & 0.962 & 0.949 & 29.82 & 0.941 & 13.69 \\
 & 4K & 25.70 & 0.856 & 6.58 & 0.896 & 0.934 & 0.954 & 0.938 & 29.80 & 0.940 & 13.62 \\
 & 5K & 26.56 & 0.870 & 6.33 & 0.913 & 0.948 & 0.958 & 0.949 & 30.56 & 0.942 & 14.26 \\
 & 10K & 26.44 & 0.864 & 7.16 & 0.914 & 0.952 & 0.956 & 0.951 & 30.10 & 0.940 & 13.28 \\
 & 18K & \textbf{27.23} & 0.879 & 6.18 & 0.919 & 0.952 & 0.962 & \textbf{0.954} & \textbf{31.35} & \textbf{0.950} & 13.21 \\
 & 20K & 27.16 & \textbf{0.880} & \textbf{5.97} & 0.919 & \textbf{0.971} & 0.962 & 0.953 & 30.82 & 0.948 & 12.00 \\
 & 30K & 26.60 & 0.873 & 6.30 & 0.912 & 0.947 & 0.959 & 0.949 & 30.30 & 0.947 & \textbf{10.35} \\
 & 40K & 26.43 & 0.867 & 6.63 & 0.911 & 0.948 & 0.956 & 0.948 & 30.40 & 0.944 & 14.34 \\
 & 50K & 26.82 & 0.875 & 6.23 & \textbf{0.920} & 0.953 & 0.960 & \textbf{0.954} & 30.29 & 0.949 & 10.93 \\
 & 100K & 26.84 & 0.873 & 6.32 & 0.918 & 0.950 & 0.962 & 0.953 & 30.63 & 0.945 & 11.91 \\
 & 500K & 26.75 & 0.873 & 6.12 & 0.916 & 0.949 & \textbf{0.963} & 0.953 & 30.89 & 0.947 & 12.45 \\
\bottomrule
\end{tabular}}
\end{table}

\begin{table}[t]
\centering
\small
\caption{Quantitative comparison between Qwen-Image-Layered~\citep{yin2025qwen}, the original PrismLayersPro trained CLD baseline, and our best medium-scale SynLayers-trained checkpoint.}
\label{tab:qwen_cld_qual}
\begin{tabular}{l|ccc|ccc}
\toprule
\multirow{2}{*}{Model} & \multicolumn{3}{c|}{Layer-wise} & \multicolumn{3}{c}{Composite} \\
\cline{2-7}
& PSNR $\uparrow$ & SSIM $\uparrow$ & FID $\downarrow$ & PSNR $\uparrow$ & SSIM $\uparrow$ & FID $\downarrow$ \\
\midrule
Qwen-Image-Layered~\citep{yin2025qwen} & 13.80 & 0.509 & 198.34 & 27.19 & 0.899 & 33.23 \\
PrismLayersPro & 26.22 & 0.865 & 6.62 & 30.52 & 0.944 & 12.50 \\
\textbf{SynLayers (Ours)} & \textbf{27.16} & \textbf{0.880} & \textbf{5.97} & \textbf{30.82} & \textbf{0.948} & \textbf{12.00} \\
\bottomrule
\end{tabular}
\end{table}

\begin{table}[t]
    \centering
    \small
    \caption{Composite-only evaluation on our 147-image out-of-distribution real-world testset. For the CLD-based methods, inference inputs are produced automatically by our trained Qwen3-VL-based detector, which predicts the whole-image caption and layer bounding boxes for each image.}
    \label{tab:ood_real_eval}
    \begin{tabular}{lccc}
    \toprule
    Model & PSNR $\uparrow$ & SSIM $\uparrow$ & FID $\downarrow$ \\
    \midrule
    Qwen-Image-Layered~\citep{yin2025qwen} & 28.56 & 0.843 & 64.30 \\
    PrismLayersPro & 28.74 & 0.852 & 44.23 \\
    \textbf{SynLayers (Ours)} & \textbf{29.35} & \textbf{0.856} & \textbf{35.40} \\
    \bottomrule
\end{tabular}
\end{table}

We further assess the generalization ability among different models on a 147-image out-of-distribution (OOD) real-world test set, where inference is fully automated by our fine-tuned Qwen3-VL detector, which can predict image-level captions and layer bounding boxes. Detailed detector evaluations are in Appendix~\ref{app:detector_evaluation}. As layer-level ground truth is unavailable, we report composite reconstruction metrics in Table~\ref{tab:ood_real_eval}. Our SynLayers-trained checkpoint consistently outperforms the original baseline CLD, such as PSNR from $28.74$ to $29.35$ and FID from $44.23$ to $35.40$, demonstrating robust transfer to diverse unseen layouts. In contrast, Qwen-Image-Layed~\citep{yin2025qwen} exhibits a sharp increase in FID to $64.30$, reflecting weaker distribution-level realism under OOD conditions. These results provide preliminary evidence that SynLayers can achieve superior performance in real-world layer-decomposition tasks. Revealing the potential to act as the foundational method of constructing a dataset in image editing and layer design domains.

\subsection{Discussion}

\begin{figure*}[!htbp]
    \centering
    \begin{subfigure}[t]{0.247\textwidth}
        \centering
        \begin{tikzpicture}
            \begin{axis}[
                width=1.25\linewidth,
                height=0.9\linewidth,
                title={PSNR},
                title style={font=\small},
                xmin=10, xmax=500,
                ymin=24.3, ymax=31.2,
                xtick={10,50,100,200,300,450,500},
                grid=major,
                unbounded coords=jump,
                tick label style={font=\tiny},
                x tick label style={rotate=45, anchor=east, font=\tiny},
                every axis plot/.append style={line width=1.0pt, mark size=1.9pt},
                legend to name=trainingdynamicslegendshared,
                legend style={
                    font=\scriptsize,
                    draw=none,
                    fill=none,
                    legend columns=4,
                    /tikz/every even column/.append style={column sep=0.7em}
                },
                clip=false,
            ]
            \addplot+[color=blue, mark=*] coordinates {
                (10,25.533) (20,26.110) (30,26.268) (40,26.755) (50,26.810) (60,27.028) (80,26.009) (90,26.048)
                (100,26.260) (150,26.590) (200,27.012) (250,26.765) (300,26.928) (350,27.012) (450,26.846) (500,26.738)
            };
            \addplot+[color=red, mark=square*] coordinates {
                (10,29.238) (20,29.898) (30,30.124) (40,30.889) (50,30.162) (60,30.830) (80,29.509) (90,29.659)
                (100,30.060) (150,30.251) (200,30.852) (250,30.577) (300,30.847) (350,30.940) (450,30.902) (500,30.784)
            };
            \addlegendimage{color=blue, mark=*, line width=1.0pt}
            \addlegendentry{Layer PSNR}
            \addlegendimage{color=red, mark=square*, line width=1.0pt}
            \addlegendentry{Composite PSNR}
            \addlegendimage{color=teal, mark=triangle*, line width=1.0pt}
            \addlegendentry{Layer SSIM}
            \addlegendimage{color=orange, mark=diamond*, line width=1.0pt}
            \addlegendentry{Composite SSIM}
            \addlegendimage{color=black, mark=x, line width=1.0pt}
            \addlegendentry{Layer FID}
            \addlegendimage{color=magenta, mark=+, line width=1.0pt}
            \addlegendentry{Composite FID}
            \addlegendimage{color=violet, mark=pentagon*, line width=1.0pt}
            \addlegendentry{Mask IoU}
            \addlegendimage{color=ForestGreen, mark=star, line width=1.0pt}
            \addlegendentry{Mask F1}
            \end{axis}
        \end{tikzpicture}
    \end{subfigure}\hfill
    \begin{subfigure}[t]{0.247\textwidth}
        \centering
        \begin{tikzpicture}
            \begin{axis}[
                width=1.25\linewidth,
                height=0.9\linewidth,
                title={SSIM},
                title style={font=\small},
                xmin=10, xmax=500,
                ymin=0.82, ymax=0.95,
                xtick={10,50,100,200,300,450,500},
                grid=major,
                unbounded coords=jump,
                tick label style={font=\tiny},
                x tick label style={rotate=45, anchor=east, font=\tiny},
                every axis plot/.append style={line width=1.0pt, mark size=1.9pt},
                clip=false,
            ]
            \addplot+[color=teal, mark=triangle*] coordinates {
                (10,0.852) (20,0.863) (30,0.866) (40,0.874) (50,0.875) (60,0.877) (80,0.858) (90,0.858)
                (100,0.863) (150,0.869) (200,0.876) (250,0.871) (300,0.873) (350,0.875) (450,0.872) (500,0.868)
            };
            \addplot+[color=orange, mark=diamond*] coordinates {
                (10,0.935) (20,0.939) (30,0.940) (40,0.947) (50,0.938) (60,0.945) (80,0.932) (90,0.928)
                (100,0.938) (150,0.937) (200,0.942) (250,0.940) (300,0.943) (350,0.944) (450,0.944) (500,0.943)
            };
            \end{axis}
        \end{tikzpicture}
    \end{subfigure}\hfill
    \begin{subfigure}[t]{0.247\textwidth}
        \centering
        \begin{tikzpicture}
            \begin{axis}[
                width=1.25\linewidth,
                height=0.9\linewidth,
                title={Mask Metrics},
                title style={font=\small},
                xmin=10, xmax=500,
                ymin=0.87, ymax=0.97,
                xtick={10,50,100,200,300,450,500},
                x tick label style={rotate=45, anchor=east, font=\tiny},
                grid=major,
                unbounded coords=jump,
                tick label style={font=\tiny},
                every axis plot/.append style={line width=1.0pt, mark size=1.9pt},
                clip=false,
            ]
            \addplot+[color=violet, mark=pentagon*] coordinates {
                (10,0.901) (20,0.906) (30,0.910) (40,0.916) (50,0.913) (60,0.918) (80,0.904) (90,0.905)
                (100,0.909) (150,0.912) (200,0.916) (250,0.916) (300,0.918) (350,0.918) (450,0.915) (500,0.916)
            };
            \addplot+[color=ForestGreen, mark=star] coordinates {
                (10,0.941) (20,0.945) (30,0.948) (40,0.951) (50,0.949) (60,0.953) (80,0.944) (90,0.945)
                (100,0.947) (150,0.949) (200,0.952) (250,0.952) (300,0.953) (350,0.953) (450,0.951) (500,0.952)
            };
            \end{axis}
        \end{tikzpicture}
    \end{subfigure}\hfill
    \begin{subfigure}[t]{0.247\textwidth}
        \centering
        \begin{tikzpicture}
            \begin{axis}[
                width=1.25\linewidth,
                height=0.9\linewidth,
                title={FID},
                title style={font=\small},
                xmin=10, xmax=500,
                ymin=0, ymax=40,
                xtick={10,50,100,200,300,450,500},
                x tick label style={rotate=45, anchor=east, font=\tiny},
                grid=major,
                unbounded coords=jump,
                tick label style={font=\tiny},
                every axis plot/.append style={line width=1.0pt, mark size=1.9pt},
                clip=false,
            ]
            \addplot+[color=black, mark=x] coordinates {
                (10,7.028) (20,6.442) (30,6.561) (40,6.126) (50,6.668) (60,6.178) (80,8.700) (90,8.975)
                (100,8.063) (150,7.278) (200,7.004) (250,7.420) (300,7.056) (350,7.149) (450,7.452) (500,7.412)
            };
            \addplot+[color=magenta, mark=+] coordinates {
                (10,18.267) (20,14.242) (30,15.968) (40,12.453) (50,18.249) (60,14.522) (80,22.251) (90,27.224)
                (100,20.488) (150,20.174) (200,20.568) (250,21.850) (300,20.253) (350,21.347) (450,20.216) (500,20.605)
            };
            \end{axis}
        \end{tikzpicture}
    \end{subfigure}

    \vspace{0em}

    {\centering\pgfplotslegendfromname{trainingdynamicslegendshared}\par}

    \caption{Checkpoint-level training dynamics of the full SynLayers-trained model. All curves use the recorded evaluation values at each retained checkpoint from 10K to 500K.}
    \label{fig:training_dynamics_app}
\end{figure*}
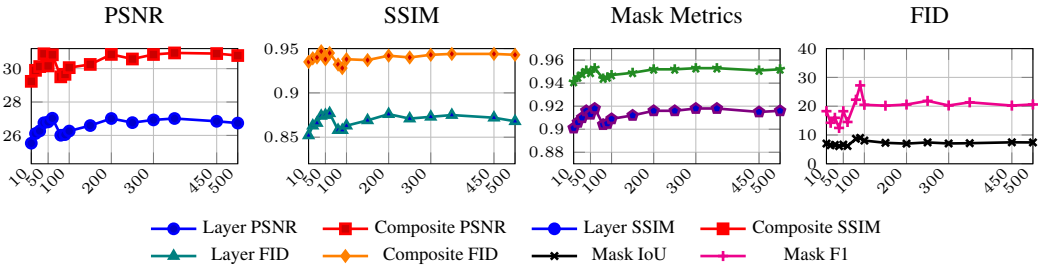

\textbf{Training dynamics of the 500K model.} Figure~\ref{fig:training_dynamics_app} tracks evaluation metrics across checkpoints from 10K to 500K. Training exhibits rapid early gains followed by an oscillatory plateau. Using Composite PSNR and FID as indicators, reconstruction quality improves sharply from 29.24 to 30.89 from 10K to 40K, then stabilizes. Conversely, distributional realism peaks at 40K with an FID of 12.45, degrades to 18.25 by 50K, and subsequently fluctuates in the 20-22 range. This establishes the 40K-60K window as the optimal trade-off, balancing pixel fidelity and distributional alignment. Extending training to 500K maintains reconstruction quality but yields no further gains.

\textbf{Performance by layer count}. A key advantage of SynLayers is the ability to construct training samples with arbitrary layer counts. The PrismLayersPro training set is constrained by the layer structure or layout design of real design files, which limits the diversity of layer counts during training. In contrast, SynLayers flexibly comprises a varying number of layers using multi-source images. To test whether that flexibility improves robustness across design complexity, we regroup the PrismLayersPro test set into four contiguous ground-truth layer-count bins chosen to be as even as possible over the 1 to 35 layer range: 1 to 7, 8 to 9, 10 to 12, and 13 to 35 layers. Each bin contains 300, 265, 247, and 187 samples, respectively. As shown in Figure \ref{fig:layercount_overview_app}, Layer-reconstruction metrics (Fig.~\ref{fig:layer_metric_app}) consistently favor training using SynLayers, with the 18K SynLayers subset trained checkpoint leading in pixelwise quality and Layer FID improving across all bins. Composite results (Fig.~\ref{fig:composite_metric_app}) are more heterogeneous: 18K achieves the highest PSNR, while 30K yields the lowest FID, reinforcing our core trade-off between reconstruction fidelity and distributional realism. Mask quality (Fig.~\ref{fig:mask_metric_app}) also consistently benefits from synthetic supervision, with 18K--50K checkpoints outperforming the baseline in IoU and F1 across all complexity levels. Overall, SynLayers can reliably enhance editable structure and mask accuracy, with performance stabilizing at moderate scales.

\pgfplotsset{
    layercountmetric/.style={
        xmin=1, xmax=4,
        xtick={1,2,3,4},
        xticklabels={1 to 7,8 to 9,10 to 12,13 to 35},
        grid=major,
        unbounded coords=jump,
        tick label style={font=\scriptsize},
        x tick label style={rotate=35, anchor=east, font=\tiny},
        title style={font=\small},
        every axis plot/.append style={line width=1.0pt, mark size=1.9pt},
        clip=false,
    }
}

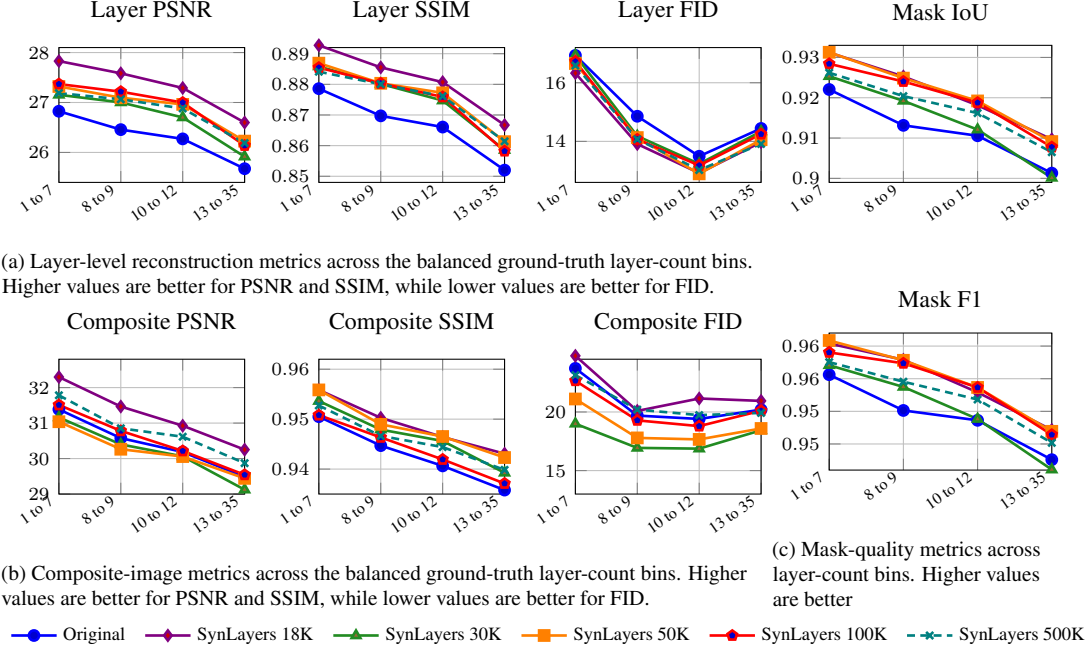
\begin{figure}[t]
    \centering
    \begin{minipage}[t]{0.73\linewidth}
        \begin{subfigure}[t]{\linewidth}
            \centering
            \begin{minipage}[t]{0.33\linewidth}
                \centering
                \begin{tikzpicture}
                    \begin{axis}[
                        layercountmetric,
                        width=1.2\linewidth,
                        height=\linewidth,
                        title={Layer PSNR},
                        ymin=25.4, ymax=28.1,
                        legend to name=layercountlegendshared,
                        legend columns=6,
                        legend style={
                            font=\scriptsize,
                            draw=none,
                            fill=none,
                            /tikz/every even column/.append style={column sep=0.5em}
                        },
                    ]
                        \addplot+[color=blue, mark=*] coordinates {(1,26.824663) (2,26.456720) (3,26.271261) (4,25.674507)};
                        \addplot+[color=violet, mark=diamond*] coordinates {(1,27.829097) (2,27.584200) (3,27.290741) (4,26.598492)};
                        \addplot+[color=ForestGreen, mark=triangle*] coordinates {(1,27.153973) (2,27.001916) (3,26.700931) (4,25.915608)};
                        \addplot+[color=orange, mark=square*] coordinates {(1,27.321007) (2,27.095507) (3,26.962055) (4,26.226737)};
                        \addplot+[color=red, mark=pentagon*] coordinates {(1,27.369340) (2,27.216652) (3,26.998787) (4,26.138717)};
                        \addplot+[color=teal, mark=x] coordinates {(1,27.188114) (2,27.063785) (3,26.880253) (4,26.188210)};
                        \legend{Original, SynLayers 18K, SynLayers 30K, SynLayers 50K, SynLayers 100K, SynLayers 500K}
                    \end{axis}
                \end{tikzpicture}
            \end{minipage}\hfill
            \begin{minipage}[t]{0.33\linewidth}
                \centering
                \begin{tikzpicture}
                    \begin{axis}[
                        layercountmetric,
                        width=1.2\linewidth,
                        height=\linewidth,
                        title={Layer SSIM},
                        ymin=0.848, ymax=0.892,
                    ]
                        \addplot+[color=blue, mark=*] coordinates {(1,0.878590) (2,0.869717) (3,0.866060) (4,0.852004)};
                        \addplot+[color=violet, mark=diamond*] coordinates {(1,0.892732) (2,0.885533) (3,0.880718) (4,0.866706)};
                        \addplot+[color=ForestGreen, mark=triangle*] coordinates {(1,0.885828) (2,0.880144) (3,0.874707) (4,0.858570)};
                        \addplot+[color=orange, mark=square*] coordinates {(1,0.886914) (2,0.880338) (3,0.877222) (4,0.861292)};
                        \addplot+[color=red, mark=pentagon*] coordinates {(1,0.885451) (2,0.880313) (3,0.875901) (4,0.858203)};
                        \addplot+[color=teal, mark=x] coordinates {(1,0.884127) (2,0.880043) (3,0.876138) (4,0.861399)};
                    \end{axis}
                \end{tikzpicture}
            \end{minipage}\hfill
            \begin{minipage}[t]{0.33\linewidth}
                \centering
                \begin{tikzpicture}
                    \begin{axis}[
                        layercountmetric,
                        width=1.2\linewidth,
                        height=\linewidth,
                        title={Layer FID},
                        ymin=12.6, ymax=17.2,
                    ]
                        \addplot+[color=blue, mark=*] coordinates {(1,16.932844) (2,14.856848) (3,13.492141) (4,14.440795)};
                        \addplot+[color=violet, mark=diamond*] coordinates {(1,16.327833) (2,13.909333) (3,12.917351) (4,13.970915)};
                        \addplot+[color=ForestGreen, mark=triangle*] coordinates {(1,16.934177) (2,14.180116) (3,13.232191) (4,14.339338)};
                        \addplot+[color=orange, mark=square*] coordinates {(1,16.656182) (2,14.129367) (3,12.891430) (4,14.057346)};
                        \addplot+[color=red, mark=pentagon*] coordinates {(1,16.725725) (2,14.068036) (3,13.178076) (4,14.247737)};
                        \addplot+[color=teal, mark=x] coordinates {(1,16.587043) (2,14.071980) (3,13.021413) (4,13.912219)};
                    \end{axis}
                \end{tikzpicture}
            \end{minipage}
            \caption{Layer-level reconstruction metrics across the balanced ground-truth layer-count bins. Higher values are better for PSNR and SSIM, while lower values are better for FID.}
            \label{fig:layer_metric_app}
        \end{subfigure}


        \begin{subfigure}[t]{\linewidth}
            \centering
            \begin{minipage}[t]{0.33\linewidth}
                \centering
                \begin{tikzpicture}
                    \begin{axis}[
                        layercountmetric,
                        width=1.2\linewidth,
                        height=\linewidth,
                        title={Composite PSNR},
                        ymin=29.0, ymax=32.8,
                    ]
                        \addplot+[color=blue, mark=*] coordinates {(1,31.386227) (2,30.576403) (3,30.185121) (4,29.483260)};
                        \addplot+[color=violet, mark=diamond*] coordinates {(1,32.297669) (2,31.466716) (3,30.925611) (4,30.251415)};
                        \addplot+[color=ForestGreen, mark=triangle*] coordinates {(1,31.160081) (2,30.399450) (3,30.061716) (4,29.123356)};
                        \addplot+[color=orange, mark=square*] coordinates {(1,31.038470) (2,30.263374) (3,30.056928) (4,29.442289)};
                        \addplot+[color=red, mark=pentagon*] coordinates {(1,31.511221) (2,30.781240) (3,30.214853) (4,29.542438)};
                        \addplot+[color=teal, mark=x] coordinates {(1,31.783633) (2,30.853348) (3,30.615276) (4,29.870712)};
                    \end{axis}
                \end{tikzpicture}
            \end{minipage}\hfill
            \begin{minipage}[t]{0.33\linewidth}
                \centering
                \begin{tikzpicture}
                    \begin{axis}[
                        layercountmetric,
                        width=1.2\linewidth,
                        height=\linewidth,
                        title={Composite SSIM},
                        ymin=0.935, ymax=0.962,
                    ]
                        \addplot+[color=blue, mark=*] coordinates {(1,0.950498) (2,0.944699) (3,0.940640) (4,0.935798)};
                        \addplot+[color=violet, mark=diamond*] coordinates {(1,0.955849) (2,0.950264) (3,0.946419) (4,0.943081)};
                        \addplot+[color=ForestGreen, mark=triangle*] coordinates {(1,0.953630) (2,0.947909) (3,0.945627) (4,0.939256)};
                        \addplot+[color=orange, mark=square*] coordinates {(1,0.955854) (2,0.948963) (3,0.946522) (4,0.942312)};
                        \addplot+[color=red, mark=pentagon*] coordinates {(1,0.950808) (2,0.946346) (3,0.941928) (4,0.937116)};
                        \addplot+[color=teal, mark=x] coordinates {(1,0.952739) (2,0.946643) (3,0.944406) (4,0.939840)};
                    \end{axis}
                \end{tikzpicture}
            \end{minipage}\hfill
            \begin{minipage}[t]{0.33\linewidth}
                \centering
                \begin{tikzpicture}
                    \begin{axis}[
                        layercountmetric,
                        width=1.2\linewidth,
                        height=\linewidth,
                        title={Composite FID},
                        ymin=13.0, ymax=24.5,
                    ]
                        \addplot+[color=blue, mark=*] coordinates {(1,23.719456) (2,19.693415) (3,19.405109) (4,20.198048)};
                        \addplot+[color=violet, mark=diamond*] coordinates {(1,24.788227) (2,20.091461) (3,21.150074) (4,20.944713)};
                        \addplot+[color=ForestGreen, mark=triangle*] coordinates {(1,18.996409) (2,16.936752) (3,16.872804) (4,18.440241)};
                        \addplot+[color=orange, mark=square*] coordinates {(1,21.108496) (2,17.788751) (3,17.660815) (4,18.593099)};
                        \addplot+[color=red, mark=pentagon*] coordinates {(1,22.663849) (2,19.285285) (3,18.793016) (4,20.106108)};
                        \addplot+[color=teal, mark=x] coordinates {(1,23.143214) (2,20.204466) (3,19.730423) (4,19.921571)};
                    \end{axis}
                \end{tikzpicture}
            \end{minipage}
            \caption{Composite-image metrics across the balanced ground-truth layer-count bins. Higher values are better for PSNR and SSIM, while lower values are better for FID.}
            \label{fig:composite_metric_app}
        \end{subfigure}
    \end{minipage}\hfill
    \begin{minipage}[t]{0.27\linewidth}
        \begin{subfigure}[t]{\linewidth}
            \centering
            \begin{tikzpicture}
                \begin{axis}[
                    layercountmetric,
                    width=1.2\linewidth,
                    height=0.9\linewidth,
                    title={Mask IoU},
                    ymin=0.899, ymax=0.933,
                ]
                    \addplot+[color=blue, mark=*] coordinates {(1,0.922049) (2,0.913153) (3,0.910571) (4,0.901300)};
                    \addplot+[color=violet, mark=diamond*] coordinates {(1,0.931206) (2,0.925270) (3,0.918122) (4,0.909686)};
                    \addplot+[color=ForestGreen, mark=triangle*] coordinates {(1,0.925339) (2,0.919230) (3,0.912089) (4,0.900129)};
                    \addplot+[color=orange, mark=square*] coordinates {(1,0.931277) (2,0.924868) (3,0.919209) (4,0.909128)};
                    \addplot+[color=red, mark=pentagon*] coordinates {(1,0.928421) (2,0.924032) (3,0.918778) (4,0.907759)};
                    \addplot+[color=teal, mark=x] coordinates {(1,0.926204) (2,0.920407) (3,0.916189) (4,0.906494)};
                \end{axis}
            \end{tikzpicture}

            \vspace{1.7em}

            \begin{tikzpicture}
                \begin{axis}[
                    layercountmetric,
                    width=1.2\linewidth,
                    height=0.9\linewidth,
                    title={Mask F1},
                    ymin=0.941, ymax=0.962,
                ]
                    \addplot+[color=blue, mark=*] coordinates {(1,0.955618) (2,0.950136) (3,0.948626) (4,0.942591)};
                    \addplot+[color=violet, mark=diamond*] coordinates {(1,0.960395) (2,0.957894) (3,0.952910) (4,0.947154)};
                    \addplot+[color=ForestGreen, mark=triangle*] coordinates {(1,0.957066) (2,0.953741) (3,0.948821) (4,0.941072)};
                    \addplot+[color=orange, mark=square*] coordinates {(1,0.960856) (2,0.957823) (3,0.953663) (4,0.946939)};
                    \addplot+[color=red, mark=pentagon*] coordinates {(1,0.959028) (2,0.957363) (3,0.953665) (4,0.946420)};
                    \addplot+[color=teal, mark=x] coordinates {(1,0.957544) (2,0.954540) (3,0.951815) (4,0.945173)};
                \end{axis}
            \end{tikzpicture}
            \caption{Mask-quality metrics across layer-count bins. Higher values are better}
            \label{fig:mask_metric_app}
        \end{subfigure}
    \end{minipage}


    {\centering\pgfplotslegendfromname{layercountlegendshared}\par}
    \caption{Left column reports layer-wise and composite metrics, and the right reports mask-quality metrics.}
    \label{fig:layercount_overview_app}
\end{figure}

\begin{table}[t]
    \centering
    \small
    \caption{Layer-count distribution across the PrismLayersPro and SynLayers of different scales.}
    \label{tab:layer_statistics_app}
    \resizebox{\linewidth}{!}{
    \begin{tabular}{lrrrrrr}
    \toprule
    Layer bin / summary & Original (18K) & SynLayers 18K & SynLayers 30K & SynLayers 50K & SynLayers 100K & SynLayers 500K \\
    \midrule
    1 to 5   & 3,367 & 1,407 & 2,304 & 3,896 & 7,769 & 38,993 \\
    6 to 10  & 8,923 & 6,331 & 10,738 & 17,841 & 35,483 & 178,181 \\
    11 to 15 & 4,376 & 6,809 & 11,342 & 18,867 & 37,597 & 187,444 \\
    16 to 20 & 959   & 2,615 & 4,182 & 7,015 & 14,391 & 71,604 \\
    21 to 25 & 249   & 594   & 1,034 & 1,695 & 3,385 & 16,899 \\
    26 to 52 & 126   & 244   & 400   & 686   & 1,375 & 6,879 \\
    \midrule
    6 to 15 share & 73.9\% & 73.0\% & 73.6\% & 73.4\% & 73.1\% & 73.1\% \\
    1 to 20 share & 97.9\% & 95.3\% & 95.2\% & 95.2\% & 95.2\% & 95.2\% \\
    \bottomrule
    \end{tabular}}
\end{table}

Overall, this balanced-bin evaluation indicates that the benefit of synthetic layered design data is not confined to a narrow layer-count band. SynLayers-trained models improve most decomposition-oriented metrics across all four bins, while Table~\ref{tab:layer_statistics_app} shows that the training sample distribution of SynLayers remains concentrated in moderate-complexity samples rather than becoming uniform over all layer counts.

\section{Conclusion}

In this paper, we propose SynLayers, a synthetic layered designed dataset. Our work demonstrates that synthetic layered designed data provides a practical and scalable foundation for graphic design decomposition. By integrating a VLM-based detector with the CLD backbone, we automate inference inputs and validate that synthetic data supervision consistently enhances layer reconstruction quality and Composite image visual effect. Future work will focus on improving the grounding for irregular elements, incorporating professional blending effects, disentangling supervision from conditioning, and evaluating decomposed outputs in interactive workflows. We hope that this study encourages the broader adoption of synthetic data as a scalable foundation for controllable design systems.


{
\small
\bibliographystyle{plainnat}
\bibliography{references}
}


\clearpage
\appendix
\section{Supplementary Material}

\subsection{Additional Methodology Details}
\label{app:method_details}

\paragraph{Inherited CLD backbone.}
We keep the CLD architecture~\citep{liu2025controllable} unchanged, modifying only the supervision and conditioning pipeline. LD-DiT processes multi-layer latents by cropping each layer to its quantized bounding box $B_i^q$, flattening the tokens, and concatenating them into a single sequence:
\[
X_\text{input} = \left[\hat{z}_\text{comp}^{\text{crop}};\; \hat{z}_0^{\text{crop}};\; \ldots;\; \hat{z}_{N-1}^{\text{crop}}\right],
\]
where $\hat{z}_i^{\text{crop}} = \operatorname{Flatten}(\operatorname{Crop}(\hat{z}_i, B_i^q))$. This factorization restricts computation to valid layer supports, enabling efficient handling of variable layer counts. MLCA injects image-conditioned residuals through zero-initialized adapter projections:
\[
h_\text{block}^{(k)} \leftarrow h_\text{block}^{(k)} + \lambda \cdot \mathcal{A}_k\!\left(h_\text{MLCA}^{(k)}\right),
\]
with conditioning scale $\lambda = 0.9$. A transparent VAE decoder reconstructs RGBA outputs by scattering decoded tokens back to their spatial supports. During inference, we retain CLD's dual-condition classifier-free guidance:
\[
\hat{v} = v_\theta(x_t, t, \varnothing, I) + s \cdot \left[v_\theta(x_t, t, C_\text{text}, I) - v_\theta(x_t, t, \varnothing, I)\right],
\]
where the image $I$ is preserved in both branches as a structural anchor and only the text condition is dropped in the unconditional branch ($s = 4.0$). All backbone weights are frozen; we fine-tune only LoRA adapters (rank 64) injected into LD-DiT and MLCA.

\paragraph{Dataset construction pipeline.}
Our synthetic data generation pipeline operates on $1024 \times 1024$ RGBA canvases and follows a multi-stage composition strategy:

\begin{enumerate}
    \item \textbf{Base sample selection}: We sample from the first 18K PrismLayersPro-blended training samples (excluding the held-out test set). For each base, we retain the background canvas and randomly remove $N_{\text{remove}} \sim \mathcal{U}\{1, 4\}$ foreground layers while ensuring at least one remains.
    
    \item \textbf{Donor layer sampling}: We sample $N_{\text{donors}} \sim \mathcal{U}\{1, 4\}$ distinct donor designs and extract $N_{\text{layers}} \sim \mathcal{U}\{0, 2\}$ foreground layers from each. Donor layers preserve their original RGBA crops and bounding box dimensions.
    
    \item \textbf{Auxiliary content insertion} (probabilistic):
    \begin{itemize}
        \item \textit{LAION crops}: Added with probability $p_{\text{LAION}} = 0.60$, sized at $[0.3, 0.4] \times$ canvas dimensions.
        \item \textit{Rendered text layers}: Added with probability $p_{\text{text}} = 0.35$, sized at $[0.6, 0.8] \times$ canvas; bounding boxes are tightened to non-transparent content via $\alpha$-channel analysis.
        \item \textit{AlphaVAE foreground objects}: $N_{\text{AlphaVAE}} \sim \mathcal{U}\{0, 3\}$ objects, sized at $[0.25, 0.40] \times$ canvas, using generation prompts as captions.
    \end{itemize}
    
    \item \textbf{Overlap-minimizing placement}: For each candidate layer with dimensions $(w_j, h_j)$, we sample up to 300 random placements and select the one minimizing normalized overlap with occupied boxes:
    \[
    (x_0^*, y_0^*) = \arg\min_{(x_0, y_0) \in \mathcal{S}_j} \frac{1}{\mathrm{Area}(B_j)} \sum_{B_k \in \mathcal{B}_\text{occ}} \mathrm{Area}(B_j \cap B_k),
    \]
    where zero-overlap candidates are accepted immediately.
    
    \item \textbf{Metadata serialization}: Each sample is saved with layer-wise metadata (source type, bounding box, caption, image path) and a composite image.
\end{enumerate}

\paragraph{Caption construction and refinement.}
Raw captions are built by traversing a $3 \times 3$ spatial grid in reading order and concatenating position-tagged layer descriptions (source-specific: PrismLayersPro metadata, LAION pairings, rendered text, or AlphaVAE prompts). This grid-based draft is refined by Qwen2.5-VL-3B-Instruct, conditioned on both the composite image and raw caption, using a system prompt that requests: (1) 100 to 140 word output, (2) first ~40 words as holistic overview, (3) remaining 60 to 100 words for layer-level details, overlaps, and visible English text. More detailed illustration will be in Figure ~\ref{fig:caption_refinement_example}

\paragraph{VLM training and inference serialization.}
For auto-detector training, each synthesized sample is serialized as an instruction-following pair requesting a JSON object with keys \textbf{whole\_caption} and \textbf{boxes}. We fine-tune Qwen3-VL-8B-Instruct with LoRA (rank 8, \texttt{lora\_target=all}), cutoff length 3072, per-device batch size 4, gradient accumulation 4, learning rate $1.5 \times 10^{-4}$, cosine scheduling, warmup ratio 0.1, and 3 epochs in bf16. At inference, predicted JSON outputs will be all combined into a jsonl file, whose format can be consumed by the CLD pipeline.

\paragraph{Implementation details.}
The pipeline is implemented in Python with multiprocessing (64 workers) for scalability. Random seeds are fixed per sample for reproducibility. LAION images are capped at 20K for balanced diversity. All generated samples (500K total) are indexed in a jsonl file with deterministic ordering by sample ID.

\subsection{Caption Refinement Process Illustration}
\label{app:caption_refinement}

\begin{tcolorbox}[
    breakable,
    colback=blue!1!white,
    colframe=blue!70!black,
    title=Caption Refinement from Raw Grid-Based Description to Coherent Text,
    fonttitle=\bfseries,
    colbacktitle=blue!5!white,
    coltitle=blue!50!black,
    boxrule=1pt,
    arc=4pt,
    boxsep=5pt,
    left=6pt,
    right=6pt,
    top=6pt,
    bottom=6pt
]
\textbf{Input composite image:}

\begin{center}
    \includegraphics[width=0.5\linewidth]{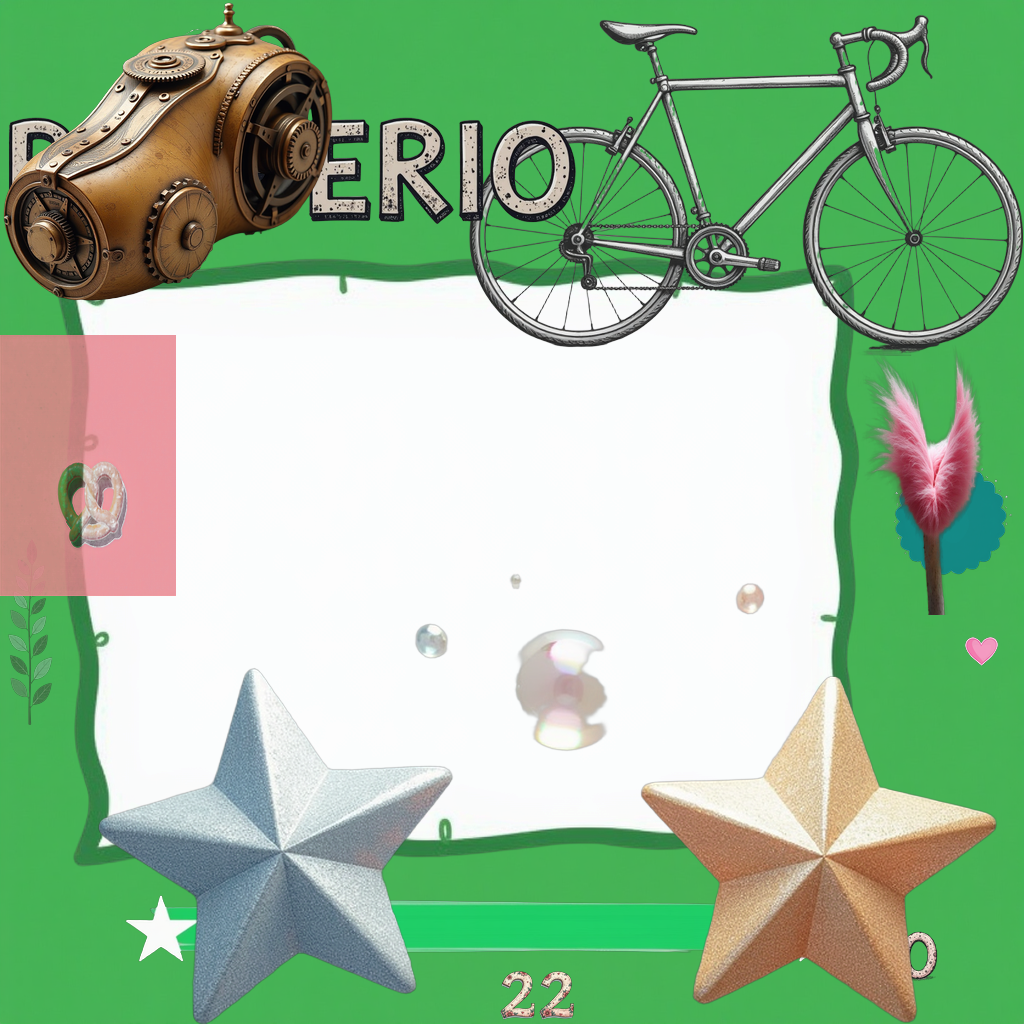}
\end{center}

\vspace{0.6em}
\textbf{Raw grid-based caption (before refinement):} \\
\small
The image you've provided is a solid, uniform color with no distinguishable features or objects. In the center, This is a doodle\_art style image. The picture shows a blank white square with a green border. On the top-right, This is a doodle art style image. A sleek bicycle is depicted with intricate line work, showcasing its frame, handlebars, and two large wheels ...... On the left, A pretzel is perfectly centered in the frame, minimalist composition, on a soft pink pastel background.

\vspace{0.6em}
\textbf{System prompt for refinement:} \\
\small
You are an expert image captioner. Your task is to refine and condense a long, redundant 'whole caption' of a layered image. Requirements: (1) Conciseness: Keep the final caption between 100 to 140 words! (2) Natural Flow: Blend the background and layers into a cohesive, professional paragraph. Avoid repetitive phrases like 'you can see' or 'there is'. (3) Output Format: Return ONLY the refined caption string. (4) Accuracy and Vividness: Ensure descriptions precisely match visual elements, using vivid but concise language. (5) First 40 words should provide an overview; remaining 60 to 100 words detail layer-level descriptions. (6) Describe overlapped layers concisely. (7) For English text layers, describe the text content in detail.

\vspace{0.6em}
\textbf{Refined caption (after Qwen2.5-VL processing):} \\
\small
A whimsical doodle art-style image features a sleek bicycle with intricate line work, a large teal circle with a playful edge, bold distressed letters spelling "RIMBERIO," and a charming heart and bubbles. The bike's minimalist yet elegant design contrasts with the playful, hand-drawn elements surrounding it.

\end{tcolorbox}

\captionof{figure}{Illustration of the caption refinement process. The raw grid-based caption (constructed by traversing the $3 \times 3$ grid and describing each layer with position tags) is refined by Qwen2.5-VL-3B-Instruct into a coherent, natural-language description. The refined caption serves as ground-truth text supervision for training both the CLD decomposition model and the Qwen3-VL detector.}
\label{fig:caption_refinement_example}

\subsection{Illustration of VLM Supervision and Inference}
\label{app:vlm_illustration}

Figure~\ref{fig:lora_qwen3vl_example} illustrates the supervision format used to fine-tune the Qwen3-VL detector and a representative post-training inference output.

\subsection{Training Details for detector}
\label{app:detector_training}

The detector, responsible for generating global captions and spatial layout constraints (bounding boxes), was trained via Supervised Fine-Tuning (SFT) on SynLayers, our spatially-annotated synthetic dataset. We fine-tuned the \texttt{Qwen3-VL-8B-Instruct} model~\citep{qwen3vl} using Low-Rank Adaptation (LoRA) within the Llama Factory framework. To ensure high-resolution spatial awareness, the input images were processed at a resolution of $1024 \times 1024$ pixels ($1{,}048{,}576$ total pixels). The model was trained on 4$\times$ NVIDIA H20 GPUs. The detailed SFT hyperparameters are summarized in Table~\ref{tab:detector_hyperparams}.

\begin{table}[!htbp]
\centering
\caption{Hyperparameters for detector LoRA Fine-tuning.}
\label{tab:detector_hyperparams}
\begin{tabular}{ll}
\toprule
\textbf{Parameter} & \textbf{Value} \\
\midrule
Base Model & \texttt{Qwen3-VL-8B-Instruct} \\
Fine-tuning Method & LoRA (Rank=8, Alpha=16) \\
LoRA Target Modules & All linear layers \\
Epochs & 3 \\
Learning Rate Scheduler & Cosine \\
Initial Learning Rate & $1.5 \times 10^{-4}$ \\
Total Batch Size & 64 \\
\quad \textit{Number of GPUs} & \textit{4 (NVIDIA H20)} \\
\quad \textit{Per-device Batch Size} & \textit{4} \\
\quad \textit{Gradient Accumulation} & \textit{4 steps} \\
Max Sequence Length & 3072 \\
Precision & \texttt{bf16} \\
Attention Mechanism & Flash-Attention 2 \\
Image Resolution & $1024 \times 1024$ \\
\bottomrule
\end{tabular}
\end{table}

\subsection{Training Details for the Decomposition Model}
\label{app:decomposition_training}

Our decomposition model is fine-tuned from \texttt{FLUX.1[dev]} ~\citep{labs2024flux}. We initialize the MultiLayer-Adapter from the released CLD adapter checkpoint and load the released CLD LoRA weights before injecting new LoRA modules into both LD-DiT and MLCA. During training, all backbone weights remain frozen except the LoRA parameters and the layer-position embeddings. The script supports both single-GPU and multi-GPU execution; in multi-GPU mode, it uses manual gradient all-reduce at accumulation boundaries instead of DDP wrappers to avoid conflicts with gradient checkpointing, and we do the training on 8 H800 GPUs. The detailed hyperparameters are summarized in Table~\ref{tab:decomposition_hyperparams}.

\begin{table}[!htbp]
\centering
\caption{Hyperparameters for decomposition-model fine-tuning.}
\label{tab:decomposition_hyperparams}
\begin{tabular}{ll}
\toprule
\textbf{Parameter} & \textbf{Value} \\
\midrule
Base Model & \texttt{FLUX.1[dev]} \\
Adapter Initialization & \texttt{FLUX.1-dev-Controlnet-Inpainting-Alpha} \\
Pretrained LoRA Initialization & Released CLD LoRA weights \\
Fine-tuning Method & LoRA on LD-DiT and MLCA \\
Trainable Parameters & LoRA weights + layer PE \\
LoRA Rank / Alpha / Dropout & 64 / 64 / 0 \\
Optimizer & Prodigy \\
Nominal Learning Rate & 1.0 \\
Betas & $(0.9, 0.999)$ \\
Weight Decay & 0.001 \\
Learning Rate Scheduler & Constant (\texttt{LambdaLR}) \\
Training Objective & Masked flow-matching MSE \\
Per-device Batch Size & 1 \\
Gradient Accumulation & 4 steps \\
Effective Batch Size & $4 \times$ number of GPUs \\
Precision & \texttt{bf16} \\
Gradient Checkpointing & Enabled for LD-DiT and MLCA \\
Max Text Sequence Length & 512 \\
Classifier-free Guidance Scale & 4.0 \\
MLCA Conditioning Scale & 0.9 \\
Inference Denoising Steps & 28 \\
Maximum Layer Number & 52 \\
Random Seed & 42 \\
\bottomrule
\end{tabular}
\end{table}

\subsection{More comparison examples between Ground Truth and SynLayers}
\label{app:gt_vs_SynLayers}
\begin{figure}[p]
    \centering
    \includegraphics[width=1\linewidth]{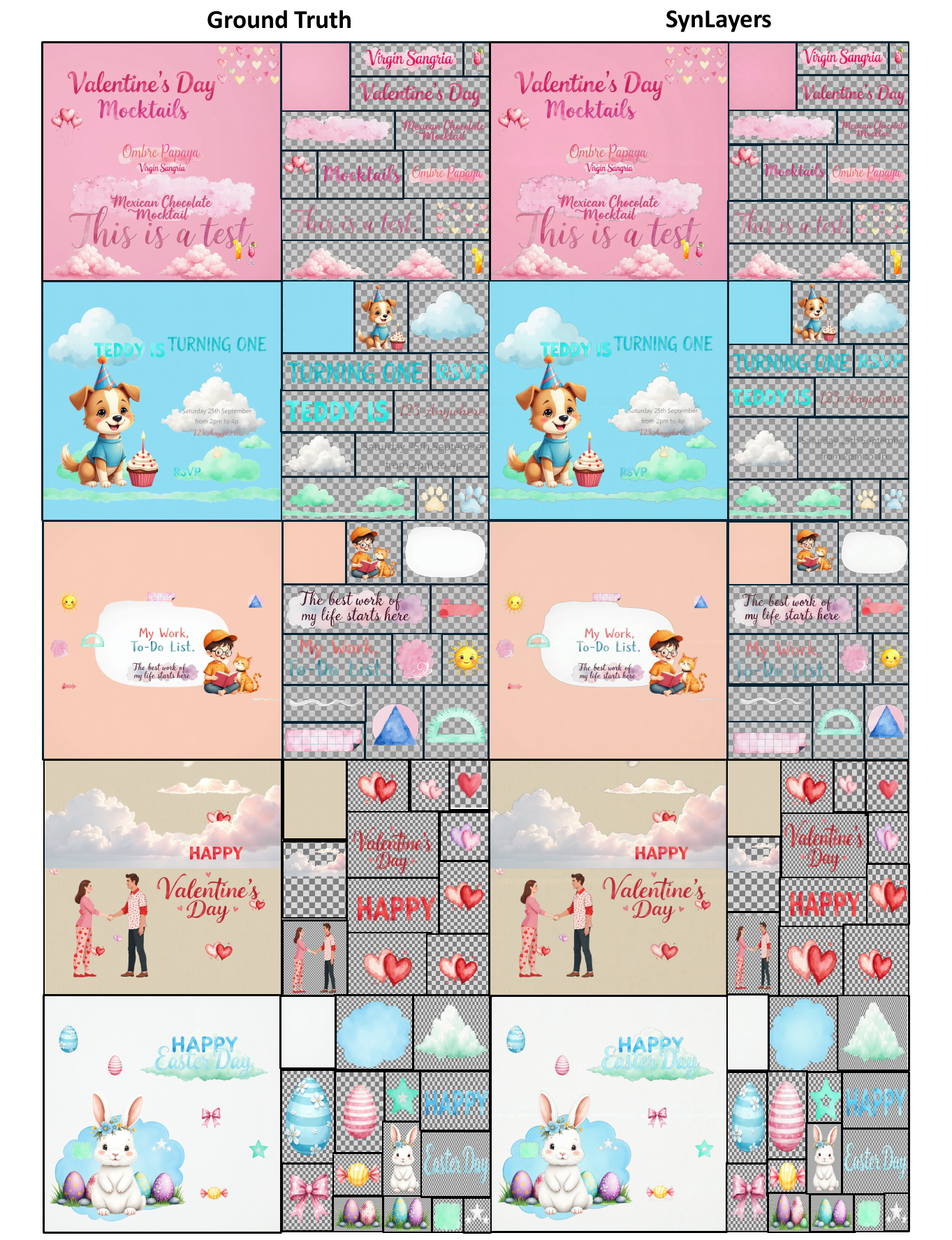}
    \caption{More comparison examples between Ground Truth and SynLayers}
    \label{fig:more_samples}
\end{figure}

In this section, we will present cases in Figure~\ref{fig:more_samples} and compare the inference output from our SynLayers-trained model with the actual ground truth.

\subsection{Quantitative Evaluation of the Qwen3-VL detector}
\label{app:detector_evaluation}

To rigorously assess the reliability of the predicted spatial constraints, Table~\ref{tab:bbox_metrics} reports both strict object-discovery metrics and matched-box localization quality for the detector. Under a strict evaluation protocol that explicitly penalizes both missed and hallucinated bounding boxes, the detector achieves 91.26\% precision, 82.34\% recall, and 86.57\% F1 at IoU$=0.50$, corresponding to 1525 true positives, 146 false positives, and 327 false negatives. Under a standard detection-style evaluation, it further obtains AP@0.50 of 76.58\%, AP@0.75 of 70.16\%, and mAP@[0.50:0.95] of 68.02\%. These results provide a substantially more realistic assessment of the detector in a fully automated pipeline than matched-box metrics alone, showing that object discovery is strong but still imperfect.

\begin{tcolorbox}[
    breakable,
    colback=red!1!white,
    colframe=red!70!black,
    title=Illustration of VLM Supervision and Inference for Automated Input Generation,
    fonttitle=\bfseries,
    colbacktitle=red!5!white,
    coltitle=red!50!black,
    boxrule=1pt,
    arc=4pt,
    boxsep=5pt,
    left=6pt,
    right=6pt,
    top=6pt,
    bottom=6pt
]
\textbf{Training sample.}

\begin{center}
    \begin{minipage}{0.47\linewidth}
        \centering
        \includegraphics[width=\linewidth]{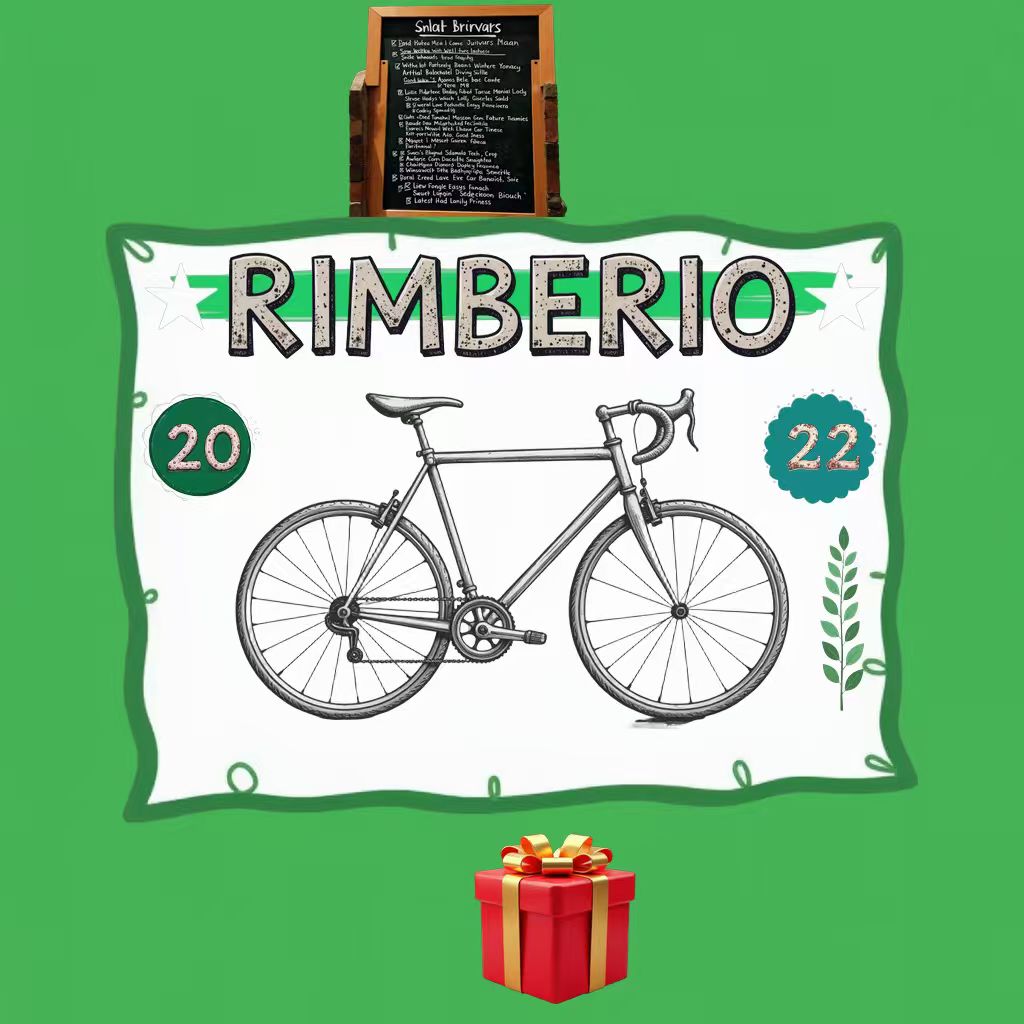}
        \captionof*{figure}{Synthetic training image}
    \end{minipage}%
    \hfill
    \begin{minipage}{0.49\linewidth}
        \small
        \textbf{Input instruction (simplified)}: \\
        \texttt{<image> This image is 1024 pixels in width and 1024 pixels in height. First describe the whole image in one detailed caption (whole\_caption). Then list the bounding box for each visible layer or object in the image. Each box is in the format [x0, y0, x1, y1]. Output a single JSON object with exactly two keys: ``whole\_caption'' and ``boxes''. Output only this JSON, no other text or markdown.}

        \vspace{0.6em}
        \textbf{Target output}: \\
        \texttt{\{"whole\_caption": "...", "boxes": [[102,212,921,825], [234,387,789,729], ...]\}}
    \end{minipage}
\end{center}

\medskip
\textbf{Inference result after LoRA fine-tuning.}

\begin{center}
    \begin{minipage}{0.47\linewidth}
        \centering
        \includegraphics[width=\linewidth]{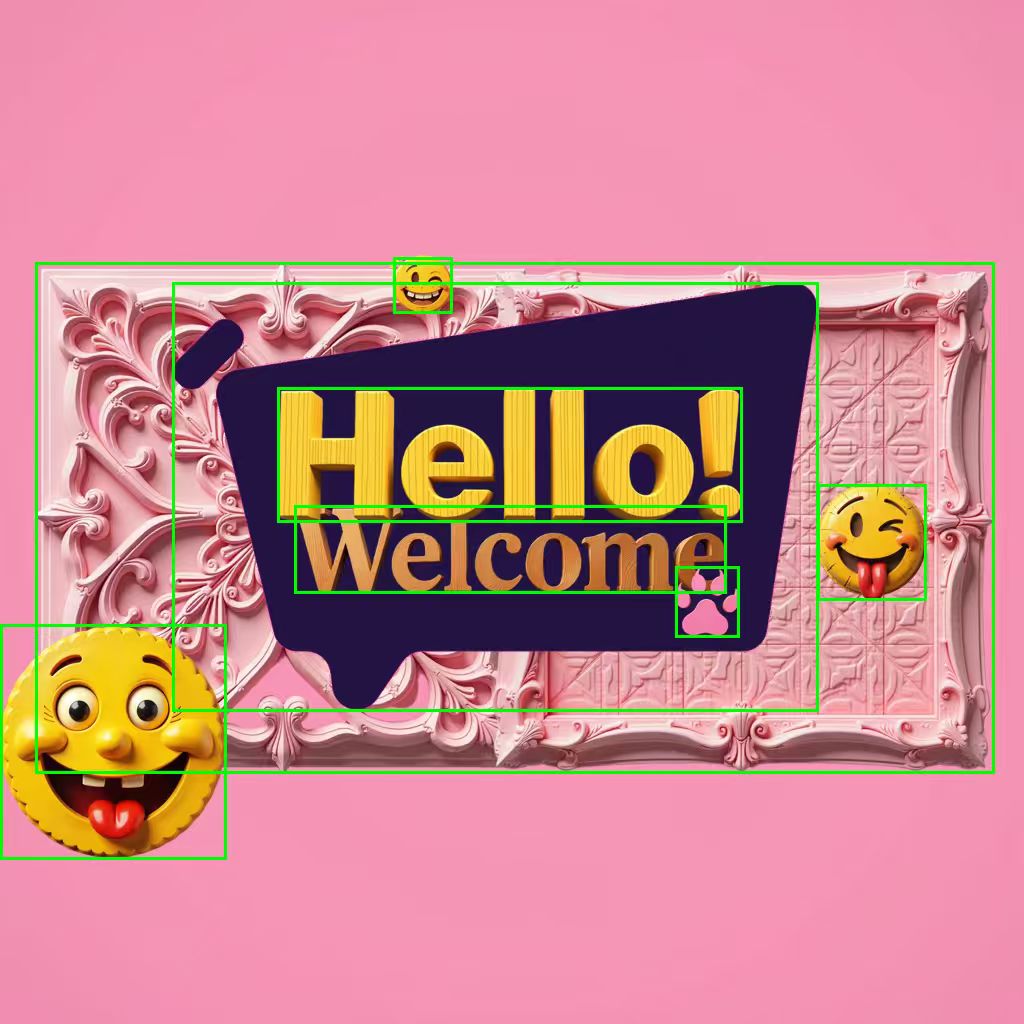}
        \captionof*{figure}{Post-training prediction with bounding-box visualization}
    \end{minipage}%
    \hfill
    \begin{minipage}{0.49\linewidth}
        \small
        \textbf{Input instruction (simplified)}: \\
        \texttt{<image> This image is 1024 pixels in width and 1024 pixels in height. First describe the whole image in one detailed caption (whole\_caption). Then list the bounding box for each visible layer or object in the image. Each box is in the format [x0, y0, x1, y1]. Output a single JSON object with exactly two keys: ``whole\_caption'' and ``boxes''. Output only this JSON, no other text or markdown.}

        \vspace{0.6em}
        \textbf{Predicted output}: \\
        \texttt{\{"whole\_caption": "The image features a vibrant pink background ... welcoming guests into a cozy space.",} \\
        \texttt{"boxes": [[35,262,994,773], [817,484,926,600], [0,624,226,859], [393,257,452,313],} \\
        \texttt{[277,387,742,522], [294,505,726,593], [675,566,739,637], [172,282,818,711]]\}}
    \end{minipage}
\end{center}

\medskip
\textbf{Inference flow}: \\

1. \textbf{Whole-image caption generation}. \\
\textbf{Step summary}: The model first generates \textbf{whole\_caption}, summarizing the semantic content, visual style, and overall composition of the design. \\

2. \textbf{Layer localization}. \\
\textbf{Step summary}: The model then predicts a sequence of layer boxes that localize the editable elements to be passed to the CLD backbone. \\

3. \textbf{Structured serialization}. \\
\textbf{Step summary}: Finally, the caption and boxes are organized into a machine-readable JSON object and converted into the jsonl format expected by the downstream inference wrapper. \\
\end{tcolorbox}

\captionof{figure}{Illustration of the supervision format and post-training inference for the Qwen3-VL detector. The upper example shows a synthetic training sample paired with a structured caption-and-box target, while the lower example shows a post-training prediction visualized with bounding boxes. Together, these examples illustrate how the model converts a raster image into the structured inputs required by the downstream decomposition pipeline.}
\label{fig:lora_qwen3vl_example}

For localization quality on successfully matched boxes, the detector still achieves a matched mIoU of 0.9547 and matched mGIoU of 0.9544. Furthermore, the mean center-distance error is only 3.66 pixels on a $1024\times1024$ canvas (0.0025 in normalized coordinates), indicating that once an object is detected, its spatial extent is estimated with very high geometric precision. Since the VLM detector does not produce calibrated per-box confidence scores, AP/mAP is computed with uniform scores over predicted boxes.

In addition to box prediction, we also evaluate the quality of the predicted \textit{whole\_caption} using GPT-4.1 as an LLM judge; the full prompt used for that judging procedure is given at the end of this subsection. As shown in Table~\ref{tab:whole_caption_metrics}, the predicted captions achieve an overall score of 80.77/100 on 200 evaluation samples. The model performs particularly well on coverage of salient content (4.340/5), reference alignment (4.280/5), and fluency and structure (4.975/5), indicating that it generally captures the main semantic content of the image and expresses it coherently. The main weakness lies in text accuracy and non-hallucination (3.460/5), suggesting that errors are primarily caused by incorrect transcription of visible text and occasional unsupported details. Overall, these results indicate that the detector provides not only reliable spatial priors for layer decomposition but also semantically strong global captions that can support downstream generation.

\begin{table}[!htbp]
    \centering
    \begin{minipage}[t]{0.48\textwidth}
        \centering
        \caption{Bounding-box evaluation results for the Qwen3-VL detector on a 200-sample test set. Strict object-discovery metrics penalize both missed and hallucinated boxes, whereas matched-box metrics evaluate only the localization quality of successfully matched predictions.}
        \label{tab:bbox_metrics}
        \small
        \begin{tabular}{l c}
            \hline
            \textbf{Metric} & \textbf{Value} \\
            \hline
            \multicolumn{2}{l}{\textit{Strict object discovery}} \\
            Precision@0.50 & 91.26\% \\
            Recall@0.50 & 82.34\% \\
            F1@0.50 & 86.57\% \\
            \hline
            \multicolumn{2}{l}{\textit{Detection-style AP/mAP}} \\
            AP@0.50 & 76.58\% \\
            AP@0.75 & 70.16\% \\
            mAP@[0.50:0.95] & 68.02\% \\
            \hline
            \multicolumn{2}{l}{\textit{Matched-box localization quality}} \\
            Matched mIoU & 0.9547 \\
            Matched mGIoU & 0.9544 \\
            Center distance (px) & 3.66 \\
            Center distance (normalized) & 0.0025 \\
            \hline
        \end{tabular}
    \end{minipage}\hfill
    \begin{minipage}[t]{0.48\textwidth}
        \centering
        \caption{GPT-4.1-based evaluation results for predicted \textit{whole\_caption} on the 200-sample test set. The overall score is a weighted average computed from five criteria.}
        \label{tab:whole_caption_metrics}
        \small
        \begin{tabular}{l c}
            \hline
            \textbf{Metric} & \textbf{Value} \\
            \hline
            Judged samples & 200 \\
            Overall score (0 to 100) & 80.77 \\
            \hline
            \multicolumn{2}{l}{\textit{Criterion means (1 to 5)}} \\
            Image faithfulness & 3.925 \\
            Coverage of salient content & 4.340 \\
            Reference alignment & 4.280 \\
            Text accuracy and non-hallucination & 3.460 \\
            Fluency and structure & 4.975 \\
            \hline
            \multicolumn{2}{l}{\textit{Verdict counts}} \\
            Excellent & 49 \\
            Good & 89 \\
            Mixed & 53 \\
            Poor & 8 \\
            Bad & 1 \\
            \hline
        \end{tabular}
    \end{minipage}
\end{table}

\paragraph{LLM Judge Evaluation Prompt.}
\label{app:llm_judge_prompt}
\begin{tcolorbox}[
    breakable,
    colback=red!1!white,
    colframe=red!70!black,
    title=Prompt for GPT-4.1-based Caption Evaluation,
    fonttitle=\bfseries,
    colbacktitle=red!5!white,
    coltitle=red!50!black,
    boxrule=1pt,
    arc=4pt,
    boxsep=5pt,
    left=6pt,
    right=6pt,
    top=6pt,
    bottom=6pt
]
\textbf{System Prompt:} \\
You are a strict but fair multimodal caption judge.

You will receive:
\begin{enumerate}
    \item the image
    \item a ground-truth caption (reference only, not an exact wording target)
    \item a predicted caption to evaluate
\end{enumerate}

\textbf{Evaluation policy:}
\begin{itemize}
    \item Use the image as the primary source of truth.
    \item Use the ground-truth caption only as a semantic reference.
    \item Do not punish harmless paraphrases or different ordering.
    \item Penalize hallucinated objects, text, numbers, colors, styles, or relations.
    \item Penalize missing major salient content.
    \item If the prediction captures the image well but differs stylistically from GT, still score it well.
    \item Many images contain rendered text. Incorrect text transcription should be penalized.
\end{itemize}
Return JSON only, with no markdown fences and no extra commentary.

\vspace{1em}
\hrule
\vspace{1em}

\textbf{User Prompt Template:} \\
You are evaluating one predicted whole-image caption.

\textbf{\#\# Scoring criteria} \\
Score each criterion from 1 to 5:

\textbf{1. image\_faithfulness} \textit{(Weight: 0.35)}
\begin{itemize}
    \item 5: strongly faithful to visible image content
    \item 3: mostly correct but with some noticeable mistakes or omissions
    \item 1: largely inconsistent with the image
\end{itemize}

\textbf{2. coverage\_of\_salient\_content} \textit{(Weight: 0.20)}
\begin{itemize}
    \item 5: covers most important objects, layout regions, and visually dominant text
    \item 3: covers only part of the key content
    \item 1: misses major salient content
\end{itemize}

\textbf{3. reference\_alignment} \textit{(Weight: 0.20)}
\begin{itemize}
    \item 5: semantically consistent with the ground-truth caption
    \item 3: partially aligned but misses important GT meaning
    \item 1: strongly conflicts with GT semantics
\end{itemize}

\textbf{4. text\_accuracy\_and\_non\_hallucination} \textit{(Weight: 0.20)}
\begin{itemize}
    \item 5: avoids unsupported details and handles visible text well
    \item 3: some speculative or inaccurate details
    \item 1: obvious hallucinations or badly incorrect text content
\end{itemize}

\textbf{5. fluency\_and\_structure} \textit{(Weight: 0.05)}
\begin{itemize}
    \item 5: clear, coherent, well-formed caption
    \item 3: understandable but awkward or verbose
    \item 1: hard to read or poorly structured
\end{itemize}

\textbf{\#\# Important instructions}
\begin{itemize}
    \item The image is the main evidence.
    \item The ground-truth caption is a reference, not a wording template.
    \item If GT itself seems slightly incomplete, do not force the prediction to copy it.
    \item Penalize hallucinations more than small wording differences.
    \item Focus on semantic correctness, salient visual coverage, and visible text accuracy.
\end{itemize}

\textbf{\#\# Ground-truth caption} \\
\{gt\_caption\}

\textbf{\#\# Predicted caption} \\
\{pred\_caption\}

\textbf{\#\# Output format} \\
Return a single JSON object with this exact schema: \\
\texttt{\{} \\
\hspace*{1.5em}\texttt{"image\_faithfulness": \{"score": 1-5, "reason": "short reason"\},} \\
\hspace*{1.5em}\texttt{"coverage\_of\_salient\_content": \{"score": 1-5, "reason": "short reason"\},} \\
\hspace*{1.5em}\texttt{"reference\_alignment": \{"score": 1-5, "reason": "short reason"\},} \\
\hspace*{1.5em}\texttt{"text\_accuracy\_and\_non\_hallucination": \{"score": 1-5, "reason": "short reason"\},} \\
\hspace*{1.5em}\texttt{"fluency\_and\_structure": \{"score": 1-5, "reason": "short reason"\},} \\
\hspace*{1.5em}\texttt{"verdict": "excellent|good|mixed|poor|bad",} \\
\hspace*{1.5em}\texttt{"error\_types": ["optional\_short\_tags"],} \\
\hspace*{1.5em}\texttt{"summary": "2-4 sentence overall explanation"} \\
\texttt{\}}
\end{tcolorbox}

\subsection{Compute Resources}
\label{app:compute_resources}

The main decomposition-model training runs were conducted on 4 NVIDIA H800 GPUs. In our training setup, obtaining one trained checkpoint for the main decomposition sweep required roughly 5 days of wall-clock time on these 4 GPUs. The detector training used 4 NVIDIA H20 GPUs, as summarized in Appendix~\ref{app:detector_training}. Beyond the checkpoints reported in the paper, the overall project also required additional exploratory runs and failed tests that are not individually listed in the main experimental tables.

\section{Broader Impacts}
\label{app:broader_impacts}

Our work studies whether synthetic layered design data can improve layered design decomposition, with the positive goal of reducing dependence on scarce proprietary layered assets and making editable-layer research more accessible. If successful, such methods can support practical creative workflows by making rasterized graphic designs easier to edit, analyze, and reuse in educational, research, and design-assistance settings.

\section{Limitations}
\label{app:limitations}

Several limitations remain. The synthetic pipeline has not yet model professional design effects, such as complex blending modes, stylized transparency, or highly irregular text layouts. The VLM-based detector still relies on rectangular boxes and JSON generation, which limits performance on non-rectangular or highly overlapping design elements, and our detector evaluation remains limited to box metrics and whole-caption quality rather than full downstream human-editing studies. The current evaluation also lacks repeated-seed uncertainty estimates and uses only a small out-of-distribution real-world set without layer-level ground truth for external validation.

\end{document}